\documentclass[11pt]{article}

\usepackage[final]{acl}

\usepackage{times}
\usepackage{latexsym}

\usepackage[T1]{fontenc}

\usepackage{adjustbox}

\newcommand{\fitformula}[1]{%
  \adjustbox{max width=.94\linewidth}{$\displaystyle #1$}%
}

\usepackage[utf8]{inputenc}

\usepackage{microtype}

\usepackage{inconsolata}

\usepackage{graphicx}
\usepackage{float}
\usepackage{algorithm}
\usepackage{algpseudocode}
\usepackage{amsmath}
\usepackage{placeins}
\usepackage{booktabs}
\usepackage{subcaption}
\usepackage[dvipsnames]{xcolor}
\usepackage{cleveref}
\usepackage{multirow}
\usepackage{pifont}
\newcommand{\cmark}{\textcolor{OliveGreen}{\ding{51}}}
\newcommand{\xmark}{\textcolor{BrickRed}{\ding{55}}}

\usepackage{cuted}

\usepackage{geometry}
\usepackage{listings}
\usepackage{caption}
\usepackage[most]{tcolorbox}

\usepackage{makecell}

\usepackage{amsmath}
\DeclareMathOperator*{\argmin}{arg\,min}

\crefname{section}{Sec.}{Sections}
\crefname{figure}{Fig.}{Figures}
\crefname{table}{Tab.}{Tables}
\crefname{equation}{Equation}{Equations}
\crefname{appendix}{Appendix}{Appendixs}

\definecolor{lightgrey}{rgb}{0.96,0.96,0.96}

\title{Compositional SVG Generation via VLM-Driven Hierarchical Semantic Parsing}

\author{
  Sehwan Park\textsuperscript{1*} \quad
  Taehoon Kim\textsuperscript{1*} \quad
  Geonhee Han\textsuperscript{1} \\[2pt]
  \bfseries
  Dohyun Kim\textsuperscript{1} \quad
  Seung Wook Kim\textsuperscript{2} \quad
  Paul Hongsuck Seo\textsuperscript{1} \\[4pt]
  \normalfont
  \textsuperscript{1}Dept. of CSE, Korea University \qquad
  \textsuperscript{2}KAIST AI \\[2pt]
  \texttt{\{shp216, kium100, rtrt505, a12s12, phseo\}@korea.ac.kr} \\
  \texttt{seungwookk@kaist.ac.kr}
}
\begin{document}
\maketitle

\begin{abstract}
While Vision-Language Models (VLMs) excel at visual reasoning, generating structured, editable Scalable Vector Graphics (SVG) remains a fundamental challenge. Existing pipelines predominantly yield flat, semantically agnostic collections of paths, where editing a single object requires manually identifying its constituent paths. To address this, we propose a VLM-driven agentic framework for semantic compositional SVG generation. Our pipeline recursively parses visual scenes into semantic and geometric hierarchies via top- down decomposition, visual grounding, and prompt-driven amodal occlusion recovery, ensuring each component is geometrically complete.  Furthermore, we introduce the Semantic SVG Benchmark with human-annotated semantic groups and novel sub-component metrics (Semantic Recall/Precision, PERE) to explicitly evaluate structural compositionality and functional editability. Experiments show that our natively predicted structures surpass the upper bounds of existing flat-generation methods in both grouping quality and editability, while maintaining state-of-the-art visual fidelity.
\end{abstract}

\section{Introduction}

\begin{figure}[t!]
    \centering
    \includegraphics[width=1.0\linewidth]{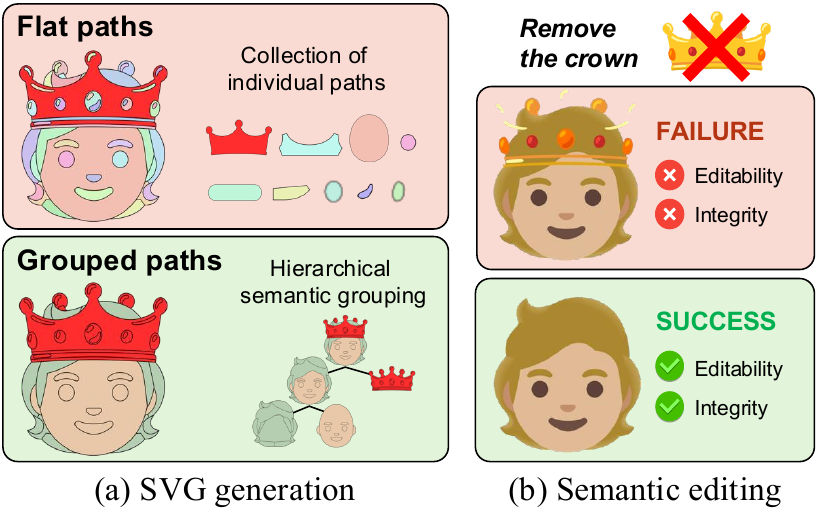}
    \caption{\textbf{The necessity of semantic compositionality in SVG generation.} (a) Prior methods output a flat, unorganized collection of paths, whereas our framework constructs semantic-aligned hierarchical tree. (b) Without a hierarchical semantic structure, downstream edits (e.g., removing the crown) fail because individual paths are decoupled from human-interpretable concepts. In contrast, our explicit semantic grouping encapsulates the entire target object into a single cohesive node, enabling successful and intuitive manipulation.}
    \label{fig:teaser}
\end{figure}

Vision-Language Models (VLMs)~\citep{anthropic2025claude37,anthropic2025claude4,bai2025qwen25vl,google2026gemini3flash,qwen2025qwen3vl,zhu2025internvl3,qwen2026qwen36} have demonstrated remarkable capabilities in bridging textual semantics with visual concepts. However, translating this high-level semantic understanding into structured, manipulable visual representations remains a fundamental challenge. Scalable Vector Graphics (SVG) serve as an ideal medium to bridge this gap, as they compose images using editable mathematical primitives. Yet, true manipulability requires these primitives to be grouped into meaningful semantic structures (e.g., organizing hundreds of disparate curves into a distinct "head" or "body" group). Without such grouping, editing a single object (e.g., moving, deleting) requires manually identifying and selecting its constituent paths, whereas encoding the semantic identity and geometric completeness of each object directly in the representation makes such edits far more reliable and efficient.

Despite the necessity of this semantic structure, existing SVG generation~\citep{duetsvg,internsvg,llm4svg,layerpeeler,starvector,omnisvg,renderingawaresvg} pipelines predominantly yield flat, semantically agnostic outputs. Whether relying on low-level optimization-based tracing~\citep{diffvg, vtracer} or direct sequence generation from VLMs~\citep{duetsvg,internsvg,starvector,omnisvg,renderingawaresvg}, current methods hit a fundamental ceiling: they generate flat vector collections that entirely miss the hierarchical semantic organization necessary for intuitive human manipulation and downstream reasoning. Recovering this organization post hoc from an already vectorized SVG~\citep{vectorprism} is likewise bounded by the given paths, which may entangle multiple objects or leave occluded regions unrecoverable.

To address this structural deficiency, we propose a paradigm shift: moving from flat SVG generation to semantic compositional SVG generation. We introduce a dynamic hierarchical parsing pipeline that leverages a multi-role VLM to recursively decompose a complex visual scene into semantically atomic sub-components. By grounding semantic concepts into pixel space, recovering occluded background geometries, and assembling the vectors into a structured markup, our approach ensures each component is geometrically complete and independently manipulable.
Furthermore, because existing evaluation protocols in recent works measure only holistic, whole-image visual fidelity, they are fundamentally incapable of assessing the internal semantic validity of the generated components. To solve this, we introduce the human-annotated Semantic SVG Benchmark together with Semantic Recall and Precision, which evaluate the quality of semantic grouping, and the Post-Edit Rendering Error (PERE), which measures functional editability by simulating a structural edit.

Our main contributions are as follows:
\begin{itemize}
    \item Semantic Compositional SVG Task and Benchmark: We formally propose the task of generating semantically grouped SVGs, and introduce the first benchmark with human-annotated semantic groups along with metrics that evaluate semantic compositionality and functional editability rather than mere whole-image fidelity.
    \item Dynamic Hierarchical Parsing Pipeline: We introduce a VLM-driven framework that parses visual scenes into compositional structures, integrating semantic grounding, targeted occlusion recovery, and assembly into a grouped SVG.
    \item State-of-the-Art Performance: Our experiments establish a rigorous baseline for this new task, showing that our pipeline outperforms existing flat-generation methods even when they are augmented with optimal post-hoc grouping.
\end{itemize}

\section{Related Work}
\paragraph{SVG Generation}
Existing SVG generation methods fall into two paradigms: optimization-based approaches~\cite{vtracer, diffvg} that reconstruct paths via low-level pixel clustering, and VLM-driven methods~\cite{llm4svg, omnisvg, internsvg, starvector, duetsvg, renderingawaresvg, layerpeeler} that predict token sequences or iteratively peel occluding layers. Both produce \emph{flat} representations, omitting structural metadata such as semantically nested \texttt{<g>} groups. Vector Prism~\cite{vectorprism} restores such grouping post hoc, yet remains bounded by the input vectorization, unable to subdivide entangled paths or recover occluded geometry. We instead emit paths only after decomposing the scene into meaningful components. This blind spot extends to evaluation: existing benchmarks~\citep{internsvg, omnisvg, starvector, llm4svg, vectorgym} rely on whole-image similarity, lacking part-level metrics. We therefore introduce a benchmark with human-annotated semantic groups, a Semantic Recall/Precision protocol, and the Post-Edit Rendering Error (PERE) measuring functional editability.

\paragraph{Tool-Augmented Multimodal Agents}
A parallel thread augments foundation models with external tools. Building on early paradigms of tool use and reasoning-action interleaving~\citep{toolformer, react} and their scaling to broader APIs and code-as-action policies~\citep{toollm, codeact}, recent work has shifted toward long-horizon, multi-tool orchestration~\citep{xu2026evolution, luo2026agentmath}. In multimodal settings, LLaVA-Plus~\citep{llavaplus} equips LMMs with vision-language tool repositories, and Visual Sketchpad~\citep{visualsketchpad} extends tool use to self-drawn artifacts, while concept-aware tools such as SAM~3~\citep{carion2026sam3segmentconcepts} make tool-augmented scene parsing practical. These systems use tools as auxiliary modules for recognition, reasoning, or short-form generation. In contrast, we use them within a structural image-to-vector process, where a single VLM assumes four cooperating roles---Hierarchical Decomposer, Residual Judge, Occlusion Assessor, and Amodal Spatial Estimator---and recursively invokes segmentation, inpainting, and vectorization tools whose outputs constitute the final SVG.

\begin{figure*}[t]
    \centering
    \includegraphics[width=1.0\linewidth]{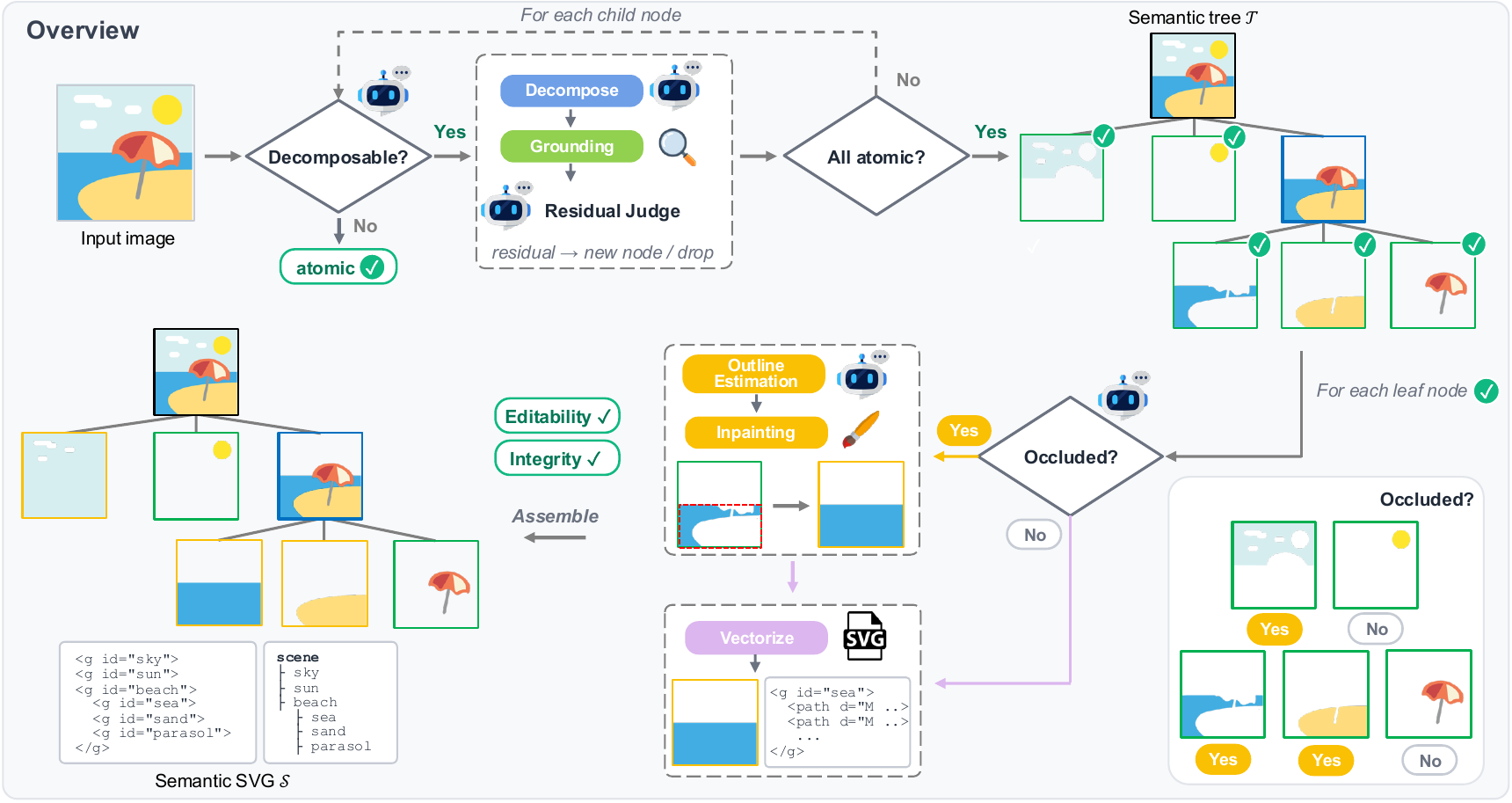}
    \caption{\textbf{Overview of our VLM-driven agentic system for compositional SVG generation.}
    (Top) Given an input image, the VLM constructs a semantic tree $\mathcal{T}$ via top-down recursion: each node is evaluated for semantic decomposability, and decomposable nodes are expanded through \emph{Decompose} (child labels and boxes in back-to-front order), \emph{Grounding} (pixel masks), and a \emph{Residual Judge} that promotes unassigned residual regions to new nodes or discards them as noise. The loop recurses on every child node until all nodes are atomic. (Bottom) For each leaf node, the VLM then judges occlusion: occluded leaves (e.g., the sea behind the sand and parasol) are amodally completed via VLM \emph{outline estimation} followed by \emph{inpainting}, while unoccluded leaves pass through unchanged. Every component is vectorized and assembled back-to-front into the semantic SVG $\mathcal{S}$, whose nested \texttt{<g>} hierarchy mirrors $\mathcal{T}$, yielding intuitively editable (\emph{Editability}) and structurally complete (\emph{Integrity}) outputs.}
    \label{fig:main_figure}
\end{figure*}

\section{Proposed Method}
\label{sec:method}

Our proposed framework transforms a raster image into a semantic, compositional SVG. Rather than generating flat vector sequences autoregressively, we formulate this as a multimodal semantic parsing problem, utilizing a multi-role Vision-Language Model (VLM) to recursively parse the visual scene into a semantic and geometric hierarchy.

\subsection{Problem Formulation: Semantic Compositional SVG Generation}
\label{sec:formulation}

Our objective is to translate a visual input $\mathcal{I}$ into a semantically structured vector representation $\mathcal{S}$. We formulate this task as the generative modeling of a hierarchical semantic tree $\mathcal{T} = (\mathcal{V}, \mathcal{E})$.

In this directed tree, vertices $\mathcal{V}$ represent discrete visual components, and edges $\mathcal{E}$ denote compositional ``part-of'' relationships. Each node $v_i \in \mathcal{V}$ is defined by a tuple $v_i = (l_i, \mathcal{S}_i)$, where $l_i$ is an explicit textual concept identifying the region. The node set $\mathcal{V}$ is partitioned into intermediate nodes and terminal leaf nodes $\mathcal{L}$. For an intermediate node, $\mathcal{S}_i$ serves as a logical grouping mechanism (mapped to an SVG \texttt{<g>} tag) that recursively encompasses its children. For a leaf node $v_k \in \mathcal{L}$, $\mathcal{S}_k$ represents a semantically atomic component, which may consist of one or more mathematical vector primitives to render its complete shape. The final SVG $\mathcal{S}$ is derived by traversing $\mathcal{T}$ and nesting the primitives according to z-order (background-to-foreground) constraints.

\subsection{Top-Down VLM-Driven Decomposition}
\label{sec:decomposition}

To construct $\mathcal{T}$, the pipeline executes a top-down recursive decomposition. The VLM acts as the core reasoning engine, converting visual complexity into a structured semantic ontology, which is then grounded into pixel space.

\noindent\textbf{Hierarchical Semantic Decomposer}\ \ \
Starting from the root image $\mathcal{I}$, we traverse the tree. Given an active node $v_i$, the VLM receives the root image and a textual prompt. Relying on its multimodal priors, it evaluates the semantic decomposability of $v_i$. If deemed decomposable, it generates a structured parsing plan: a sequence of child semantic labels $l_j$ and bounding boxes $b_j$ ordered back-to-front. If the node is evaluated as semantically atomic, it halts the descent and designates $v_i$ as a terminal leaf node $v_k \in \mathcal{L}$.

\noindent\textbf{Semantic Grounding}\ \ \
To physically execute the VLM's decomposition plan, the root image and predicted constraints $(l_j, b_j)$ are passed to the Segment Anything Model (SAM)~\cite{carion2026sam3segmentconcepts}. Here, SAM grounds the VLM's generated concepts, extracting pixel masks corresponding to the semantic labels.

\noindent\textbf{VLM Residual Judge}\ \ \
Semantic grounding rarely achieves perfect pixel mutual exclusivity. We compile any unassigned pixels into a residual image and invoke the VLM as a \emph{Residual Judge}. The VLM evaluates the residuals to determine if they constitute a missed semantic concept. If a valid semantic identity is identified, the residual is labeled and added to the tree as a new node; otherwise, it is discarded as non-semantic noise.

\noindent\textbf{Recursive Tree Expansion}\ \ \
This decomposition sequence is applied iteratively. The system dynamically expands $\mathcal{V}$ and $\mathcal{E}$ via Breadth-First Search (BFS), retaining the global image as a visual anchor while updating textual context at each depth level, continuing until all branches terminate in leaf nodes $\mathcal{L}$.

\subsection{Prompt-Driven Semantic Completion}
\label{sec:completion}

Extracting foreground components during top-down decomposition inherently leaves artificial ``holes'' in the background elements. To restore geometric integrity---which is absolutely vital to ensure the final SVG components can be independently edited by the user without revealing missing background data---the terminal leaf nodes $v_k \in \mathcal{L}$ undergo targeted occlusion recovery.

\noindent\textbf{VLM Inpainting Judge}\ \ \
Executing generative inpainting on every leaf is computationally inefficient. The VLM acts as an \emph{Occlusion Assessor}, taking the isolated leaf image, root image, and text label $l_k$ as inputs. By relying on explicitly prompted geometric priors (e.g., instructions defining the silhouette of an occluder), the VLM classifies whether true occlusion exists.

\noindent\textbf{VLM-Guided Amodal Spatial Estimation}\ \ \
For occluded leaves, the VLM functions as a \emph{Spatial Estimator}. Given the leaf image and text label, it predicts a high-resolution 2D polygon (30--50 vertices) outlining the inferred amodal (complete) shape of the component.

\noindent\textbf{Generative Inpainting}\ \ \
The target inpainting mask is derived mathematically by subtracting the visible SAM mask from the VLM-generated amodal polygon.
This exact difference mask, alongside the original leaf image and the textual label prompt, is passed to an inpainting model to recover the obscured geometry conditioned on the semantic concept $l_k$.

\subsection{Semantic Markup Assembly}
\label{sec:assembly}

Once the visual components are fully isolated and recovered, an external vectorizer translates the pixel representations of the leaf nodes $\mathcal{L}$ into mathematical primitives. Finally, the tree $\mathcal{T}$ is compiled. Because SVG is natively an XML-based markup language, we construct the final output dynamically in a bottom-up pass. We nest SVG group tags (\texttt{<g>}) using the VLM-generated semantic labels as identifiers (\texttt{id="label"}). This assembly yields a final compositional SVG where the underlying text file perfectly mirrors the semantic hierarchy extracted by the VLM.

\subsection{Extension to Text-to-SVG}
Although our framework is designed for the Image-to-SVG task, it naturally extends to Text-to-SVG by cascading a Text-to-Image (T2I) model with our pipeline. Given a text prompt $\mathcal{P}$, we synthesize a raster image $\mathcal{I}_{gen}$ via a T2I diffusion model and feed it directly into our pipeline as the root image $\mathcal{I}$. The semantic decomposition and grouping then proceed identically to the Image-to-SVG case.

\section{The Semantic SVG Benchmark}
\label{sec:benchmark}

To establish a rigorous standard for semantic compositionality in SVG generation, we introduce the \textbf{Semantic SVG Benchmark}. Because existing datasets and metrics evaluate only whole-image visual fidelity, they are incapable of measuring whether an SVG is structured for intuitive downstream manipulation. Our benchmark explicitly tests for semantic-visual alignment, geometric integrity, and structural decomposability.

\noindent\textbf{Source Data and Subsampling}\ \ \
To rigorously construct our benchmark, we collect candidate SVGs as broadly as possible from two streams. The first consists of evaluation benchmarks used in prior work~\cite{internsvg, starvector}, which span icons, complex illustrations, and emojis. The second consists of publicly available web sources, whose full list and licenses are provided in \cref{app:data-sources}.

\noindent\textbf{Rigorous Filtering Criteria}\ \ \
To guarantee the quality of the ground-truth annotations, we apply a strict manual filtering process. We retain only images with a clearly decomposable semantic structure, focusing on samples with substantial occlusion and overlapping elements so that trivial decomposition is impossible. We also exclude automatically generated samples produced by color-cluster-based optimization, whose paths span multiple semantic regions and thus make a faithful ground-truth grouping ill-defined. More details are provided in \cref{app:construction}.

\noindent\textbf{Human Annotation Process}\ \ \
Using a custom-built web interface, annotators group raw SVG paths into self-contained semantic entities (those that retain their identity when separated from the scene) and assign a textual label to each group. Starting from the root image, they recursively split each node until no meaningful decomposition remains, at which point the node is designated a leaf. More details are provided in \cref{app:construction}.

\noindent\textbf{Benchmark Statistics}\ \ \
The finalized benchmark comprises 203 rigorously annotated test samples.
The semantic trees exhibit a maximum depth of 2 and a minimum depth of 1
(representing direct children of the root), with a mean depth of 1.02.
\footnote{On average, the root node contains 2.61 ± 0.86 immediate semantic child components (median 2, range 2--8). Each image yields 2.64 ± 0.88 leaf nodes (median 2, range 2--8), and each leaf groups 7.53 ± 6.14 raw SVG paths on average (median 6, range 1--40), with the entire benchmark covering 535 leaves and 4{,}029 raw paths in total.}

\noindent\textbf{Inter-Annotator Agreement}\ \ \
To verify the reliability of our annotations, each annotator additionally labeled the samples originally annotated by the other, yielding two independent semantic trees per sample. We measure the agreement between the two trees using our grouping-quality metrics, obtaining 0.0142 (MSE) and 0.9666 (DINO), which indicates near-complete agreement. The annotators agreed on the semantic structure in most cases, and disagreements were confined to whether a few individual paths, such as background or shadow elements, belong to a given group.

\subsection{Evaluating Semantic Grouping}
\label{sec:eval_grouping}

Let $\mathcal{G} = \{g_1, \dots, g_N\}$ denote the set of non-root GT nodes, and let $\mathcal{R}(\cdot)$ be the rasterization operator that renders a set of paths into an image. We score the agreement between two rendered regions with $s(\cdot,\cdot)$, instantiated as either pixel-level MSE (lower is better) or DINO similarity (higher is better), computed over the union bounding box of the two renderings.
Throughout, $\operatorname*{best}$ denotes optimization of $s$ over
the specified domain in the metric-appropriate direction: minimization
for MSE and maximization for DINO similarity.

Given a generated SVG, let $\mathcal{Q}$ denote its complete set of
drawable paths, with $|\mathcal{Q}|=K$. Each path in $\mathcal{Q}$ is
treated as an atomic unit of grouping, and a candidate group
$p \subseteq \mathcal{Q}$ is formed by jointly rendering a subset of
these paths as a single component. We denote by $\mathcal{P}$ the set
of candidate groups against which GT nodes are matched.

For methods that natively output semantic groups
(\textit{predicted}), $\mathcal{P}=\{p_1,\dots,p_M\}$ is simply the
set of predicted groups, excluding the root. Flat baselines, however,
provide no such set. In this case (\textit{optimal}), for each GT node
$g_i$, we select the subset of generated paths that best reconstructs
it:
\begin{equation}
p_i^{\star}
=
\argmin_{p \subseteq \mathcal{Q}}
\mathrm{MSE}\bigl(\mathcal{R}(p),\mathcal{R}(g_i)\bigr).
\label{eq:optimal_group}
\end{equation}
We then use $p_i^{\star}$ as the representative group of that baseline
for $g_i$, yielding
$\mathcal{P}=\{p_1^{\star},\dots,p_N^{\star}\}$.
The exact solution defines the best GT-conditioned group that can be
formed from the baseline's paths and therefore an upper bound on its
grouping quality. In practice, we approximate this combinatorial search
using the greedy algorithm detailed in \cref{app:eval_details}.

\noindent\textbf{Semantic Recall}\ \ \
We evaluate how well the GT semantic concepts are captured by the generated output, by scoring each GT node against the candidate group that best reconstructs it:
\begin{equation}
\fitformula{
\mathrm{Recall} = \frac{1}{|\mathcal{G}|} \sum_{g_i \in \mathcal{G}} \; \operatorname*{best}_{p_j \in \mathcal{P}} \; s\bigl(\mathcal{R}(p_j), \mathcal{R}(g_i)\bigr).
}
\label{eq:recall}
\end{equation}

\noindent\textbf{Semantic Precision}\ \ \
Conversely, we evaluate the validity of the generated groups by scoring each of them against the GT node it best reconstructs, which penalizes arbitrary or semantically meaningless clusters:
\begin{equation}
\fitformula{
\mathrm{Precision} = \frac{1}{|\mathcal{P}|} \sum_{p_j \in \mathcal{P}} \; \operatorname*{best}_{g_i \in \mathcal{G}} \; s\bigl(\mathcal{R}(p_j), \mathcal{R}(g_i)\bigr).
}
\label{eq:precision}
\end{equation}
Since $\mathcal{P}$ is constructed per GT node in the \textit{optimal} setting, Precision is evaluated only in the \textit{predicted} setting, where groups are produced independently of the GT.

\subsection{Evaluating Editability and Geometric Integrity}
\label{sec:eval_editability}

Merely predicting a grouped structure is insufficient; if those groups cannot be seamlessly manipulated by a user, the grouping itself is not useful. To measure this functional editability, we introduce the Post-Edit Rendering Error (PERE), which simulates the most basic editing operation, deleting an object, and examines what the edit leaves behind. A faithful deletion requires two properties at once: the removed group must contain exactly the paths of that object, and the geometry it had occluded must be recovered so that no hole is exposed.

Let $\mathcal{E}(\mathcal{S}, g)$ denote the image obtained by deleting group $g$ from an SVG $\mathcal{S}$ and rendering the result. Given a GT sample $\mathcal{S}$ with generated counterpart $\hat{\mathcal{S}}$, we enumerate the pairs of depth-1 GT groups whose amodal masks intersect, and take the front object of each pair as the deletion target, yielding a set of GT occluders $\mathcal{O}$. For each $o \in \mathcal{O}$, we locate its counterpart in the generated SVG by rendering every candidate group in isolation and taking the closest match,
\begin{equation}
\hat{o} = \argmin_{p \in \mathcal{P}} \; \mathrm{MSE}\bigl(\mathcal{R}(p), \mathcal{R}(o)\bigr),
\label{eq:pere_match}
\end{equation}
and then delete the two counterparts and compare the resulting images over the full frame:
\begin{equation}
\mathrm{PERE} = \frac{1}{|\mathcal{O}|} \sum_{o \in \mathcal{O}} \mathrm{MSE}\bigl(\mathcal{E}(\hat{\mathcal{S}}, \hat{o}), \; \mathcal{E}(\mathcal{S}, o)\bigr),
\label{eq:pere}
\end{equation}
which we then average over the benchmark. Here $\mathcal{P}$ is the set of predicted groups for our method, and is defined as in \cref{eq:optimal_group} for flat baselines. A high PERE indicates a practical failure in compositionality: the model either grouped unrelated paths together, causing unintended elements to disappear, or failed to recover the occluded geometry, exposing a hole where the object had been. Further details are provided in \cref{app:eval_details}.

\section{Experiments}

\subsection{Experimental Settings}

\noindent\textbf{Baselines}\ \ 
For the Image-to-SVG task, we compare our method against dedicated VLM-based SVG generation models (OmniSVG-8B~\cite{omnisvg}, InternSVG-8B~\cite{internsvg}, StarVector~\cite{starvector}, LayerPeeler~\cite{layerpeeler}), a representative optimization-based baseline (VTracer~\cite{vtracer}), and general-purpose VLMs, namely open-source Qwen3.6-35B-A3B~\cite{qwen2026qwen36} and proprietary Gemini-3-flash~\cite{google2026gemini3flash}, both prompted with a dedicated instruction for semantic \texttt{<g>} grouping. For the Text-to-SVG task, we extend our pipeline with FLUX.1-dev~\cite{fluxdev} and SD3.5-medium~\cite{sd3.5} as the text-to-image models, fine-tuned with LoRA~\cite{lora} on our target domain, and compare against OmniSVG-8B, InternSVG-8B, and the same two general-purpose VLMs under the identical grouping prompts.Training details for the text-to-image models are provided in \cref{app:train_details}.

\noindent\textbf{Evaluation}\ \ 
We evaluate
Image-to-SVG models on the Semantic SVG Benchmark using Semantic
Recall and Precision (MSE and DINO), Post-Edit Rendering Error (PERE) and whole-image fidelity (MSE and
DINO). In the \textit{optimal} setting,
flat baselines are evaluated using GT-conditioned post-hoc path
selection, approximated by the greedy search detailed in
\cref{app:eval_details}. For Text-to-SVG, we evaluate generation quality and text
alignment on MMSVG-Bench~\cite{omnisvg} using FID, CLIP, HPSv2, and
Aesthetic scores.

\noindent\textbf{Implementation Details}\ \ \
We employ Gemini-3-flash~\cite{google2026gemini3flash} as the default VLM agent. For external vision tools, we use SAM~3~\cite{carion2026sam3segmentconcepts} for semantic grounding, Flux-Fill~\cite{fluxfill} for occlusion recovery, and VTracer~\cite{vtracer} for vectorization. All experiments are conducted on two NVIDIA Blackwell B200 GPUs. More implementation details are in \cref{app:implementation-details}.

\subsection{Results}
\label{sec:comparisons_to_exist_methods}
\begin{table*}[t]
\centering
\caption{\textbf{Quantitative Results on Image-to-SVG task with grouped metrics.} The \textit{optimal} represents a theoretical upper bound via post-hoc GT matching (yielding only Recall for flat baselines), while \textit{predicted} evaluates natively generated structural metadata. Notably, our purely predicted structures surpass the optimal bounds of baselines in grouping quality and functional editability, while maintaining highly competitive whole-image visual fidelity. The best and second-best results are highlighted in \textbf{bold} and \underline{underlined}, respectively.}
\label{tab:image_to_svg}
\scalebox{0.8}{%
\begin{tabular}{lcccccccccc}
\toprule
\multirow[b]{2}{*}{Methods} & \multirow[b]{2}{*}{Grouping} & \multicolumn{3}{c}{Grouping quality (MSE)$\downarrow$} & \multicolumn{3}{c}{Grouping quality (DINO)$\uparrow$} & Editability & \multicolumn{2}{c}{Whole Image} \\
\cmidrule(lr){3-5} \cmidrule(lr){6-8} \cmidrule(lr){9-9} \cmidrule(lr){10-11}
& & Recall & Precision & F1 & Recall & Precision & F1 & PERE$\downarrow$ & MSE$\downarrow$ & DINO$\uparrow$ \\
\midrule
VTracer                            & optimal   & .0227 & N/A & N/A & .9392 & N/A & N/A & .0400 & \textbf{.0036} & \textbf{.9922} \\
OmniSVG                            & optimal   & .0679 & N/A & N/A & .8276 & N/A & N/A & .1675 & .1629 & .7853 \\
InternSVG                          & optimal   & .0531 & N/A & N/A & .8581 & N/A & N/A & .0834 & .0847 & .8390 \\
StarVector                         & optimal   & .0790 & N/A & N/A & .7730 & N/A & N/A & .0969 & .1156 & .7480 \\
LayerPeeler                        & optimal   & .0295 & N/A & N/A & .9040 & N/A & N/A & .0398 & .0339 & .9148 \\
\midrule

\multirow{2}{*}{Gemini-3-flash} & optimal   & .0389 & N/A & N/A & .8983 & N/A & N/A & .0754 & .0670 & .9139 \\
                                   & predicted & .0724 & .0904 & .0787
                                    & .8937 & \underline{.8274} & \underline{.8589} & .0921 & .0670 & .9139 \\

\multirow{2}{*}{Qwen3.6-35B-A3B}   & optimal   & .0699 & N/A & N/A & .8471 & N/A & N/A & .0991 & .1070 & .8409 \\
                                   & predicted & .0801 & \underline{.0710} & \underline{.0729} & .7677 & .7413 & .7538 & .0970 & .1070 & .8409 \\

\midrule
\multirow{2}{*}{Ours}              & optimal   & \textbf{.0091} & N/A & N/A & \textbf{.9571} & N/A & N/A & \textbf{.0198} & \underline{.0049} & \underline{.9845} \\
                                   & predicted & \underline{.0174} & \textbf{.0195} & \textbf{.0155} & \underline{.9487} & \textbf{.9426} & \textbf{.9453} & \underline{.0210} & \underline{.0049} & \underline{.9845} \\

\bottomrule
\end{tabular}%
}
\end{table*}
\begin{figure*}[t!]
    \centering
    \includegraphics[width=1.0\linewidth]{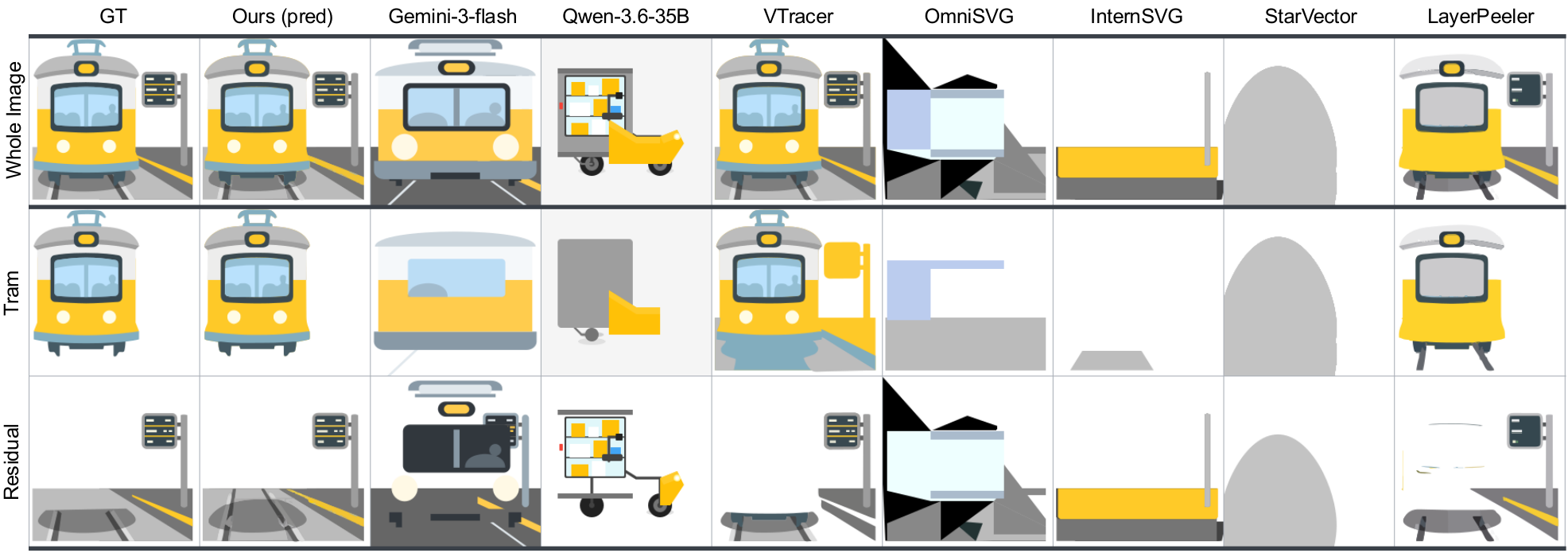}
    \caption{\textbf{Qualitative Results on the Image-to-SVG Task.} For each scenario, we show the whole image, isolated semantic object, and residual background. The whole image in the GT column is also used as the input image for all methods. Ours (pred) denotes the group (set of primitives) predicted by our model. For all remaining generated outputs, we show the semantic object rendered from paths selected using an optimal post-hoc grouping strategy. Our method effectively isolates discrete entities and recovers occluded background geometry through high-quality amodal inpainting.}
    \label{fig:qualitative_result}
\end{figure*}

\noindent\textbf{Grouping Quality}\ \ \
\cref{tab:image_to_svg} presents the quantitative results on the Image-to-SVG task, evaluated on our Semantic SVG Benchmark. Since flat generation baselines do not output semantic groups, we grant them their optimal path combination as a group so that Recall can at least be measured. While the two general-purpose VLMs do produce grouped outputs when explicitly prompted, their predicted groups fall well behind their own optimal bounds, indicating that they do not group accurately. Ours (optimal) achieves the best Recall in both MSE (0.0091) and DINO (0.9571), and Ours (predicted) relies solely on native predictions yet still surpasses the optimal configurations of all baselines across every grouping metric. This does not come at the cost of reconstruction quality: our whole-image fidelity is comparable to VTracer and far ahead of the generative baselines, while VTracer, despite its strong fidelity and the favorable optimal grouping granted to it, still falls short of our predicted groups in Recall (0.0227 vs. 0.0174 in MSE, 0.9392 vs. 0.9487 in DINO).

\noindent\textbf{Semantic Editability}\ \ \
Beyond static structure, we assess whether the generated groups are practically usable through the Post-Edit Rendering Error (PERE). Our predicted groups reduce PERE by 47.5\% relative to VTracer (0.0400 vs. 0.0210), remaining ahead of every baseline even under their optimal configurations. Flat-generation methods fail this dynamic evaluation because they merge distinct objects sharing a color into a single entangled path, or leave visible holes where an occluding component is removed. Our pipeline instead divides regions per object and recovers the occluded background through amodal inpainting, enabling clean downstream edits.

\noindent\textbf{Qualitative Results}\ \ 
\cref{fig:qualitative_result} compares the whole image, isolated semantic object, and residual background (rows) across methods (columns).
Removing the foreground tram exposes holes in the background geometry produced by VTracer, while most other baselines fail to reconstruct the target image coherently.
VTracer also merges similarly colored regions from different objects into a single path.
In contrast, our pipeline isolates components through semantic decomposition and recovers occluded geometry through amodal inpainting.
Consequently, Ours (pred) closely matches the GT in both object isolation and background completeness, showing that native predictions alone support clean edits.

\noindent\textbf{User Study on Semantic Editability}\ \ \
\begin{table}[t]
\centering
\caption{\textbf{Human editability study.} Two participants performed move and remove edits (15 samples each) on the outputs of our method and VTracer, with each edit specified by a before/after target derived from the GT SVGs. Avg.\ time is the mean time taken to complete an edit, and Completed is the number of edits finished without giving up. MSE measures the pixel-level error between the edited result and the target image from ground-truth data, computed over completed edits only. The best results are highlighted in \textbf{bold}.}
\label{tab:human_study}
\resizebox{\linewidth}{!}{%
\begin{tabular}{llccc}
\toprule
Edit & Model & Avg. time$\downarrow$ & Completed$\uparrow$ & MSE$\downarrow$ \\
\midrule
\multirow{2}{*}{Move}   & VTracer & 37.4 s          & 11/15          & 0.057 \\
                        & Ours    & \textbf{22.2 s} & \textbf{15/15} & \textbf{0.025} \\
\midrule
\multirow{2}{*}{Remove} & VTracer & 15.5 s          & 13/15          & 0.023 \\
                        & Ours    & \textbf{7.5 s}  & \textbf{15/15} & \textbf{0.006} \\
\bottomrule
\end{tabular}%
}
\end{table}
While PERE quantifies editability by simulating structural edits, we further verify whether these structures actually help users perform edits. \cref{tab:human_study} reports a user study in which participants reproduce a target edit—moving or removing objects—on the outputs of our method and VTracer. The editing environment respects the group structure of each SVG, so that selecting a group manipulates all of its paths at once, while ungrouped paths must be selected individually; participants could give up when the edit seemed impossible. All edits on our SVGs were completed, roughly $1.9\times$ faster, since an object can be manipulated as a single unit. The gain is largest for removal, where amodal inpainting restores the occluded regions and yields a substantially lower MSE (0.006 vs.\ 0.023).

\noindent\textbf{Impact of the Backbone VLM}\ \ \
\begin{table*}[t]
\centering
\caption{\textbf{Ablation on Backbone VLM.} We evaluate the impact of utilizing different VLMs, specifically Qwen3.6-35B-A3B~\cite{qwen2026qwen36}, GPT-5~\cite{openai2025gpt5}, Claude-Sonnet-4.6~\cite{anthropic2026claudesonnet46}, and Gemini-3-flash~\cite{google2026gemini3flash}, as the structural reasoning agent within our framework. Grouping quality metrics for our method are measured natively on explicitly \textit{predicted} grouping structures, whereas the baseline VTracer is evaluated using its \textit{optimal} path combinations.}
\label{tab:agent_comparison}
\scalebox{0.8}{
\begin{tabular}{lcccccccccc}
\toprule
\multirow[b]{2}{*}{Method} & \multirow[b]{2}{*}{Backbone VLM} & \multicolumn{3}{c}{Grouping quality (MSE)$\downarrow$} & \multicolumn{3}{c}{Grouping quality (DINO)$\uparrow$} & Editability & \multicolumn{2}{c}{Whole Image} \\
\cmidrule(lr){3-5} \cmidrule(lr){6-8} \cmidrule(lr){9-9} \cmidrule(lr){10-11}
& & Recall & Precision & F1 & Recall & Precision & F1 & PERE$\downarrow$ & MSE$\downarrow$ & DINO$\uparrow$ \\
\midrule

VTracer & -                & .0227 & N/A & N/A & .9392 & N/A & N/A & .0400 & \textbf{.0036} & \textbf{.9922} \\
\midrule
\multirow{4}{*}{Ours} & Qwen3.6-35B-A3B   & .0666 & .0535 & .0555 & .8541 & .8641 & .8576 & .0787 & .0123 & .9588 \\
& GPT-5             & .0629 & .0427 & .0465 & .8665 & .8945 & .8789 & .0733 & .0063 & .9771 \\
& Claude-Sonnet-4.6 & .0457 & .0417 & .0402 & .8869 & .8727 & .8788 & .0491 & .0076 & .9717 \\
& Gemini-3-flash    & \textbf{.0174} & \textbf{.0195} & \textbf{.0155} & \textbf{.9487} & \textbf{.9426} & \textbf{.9453} & \textbf{.0210} & .0049 & .9845 \\

\bottomrule
\end{tabular}%
}
\end{table*}
We evaluate the influence of the backbone VLM by comparing our default model, Gemini-3-flash, with Qwen3.6-35B-A3B, GPT-5, and Claude-Sonnet-4.6 in \cref{tab:agent_comparison}. Among the backbones, Gemini-3-flash performs best across all metrics and Claude follows, matching or exceeding most baselines in \cref{tab:image_to_svg} and remaining comparable to the strongest baseline VTracer in PERE, even though its groups are natively predicted while VTracer is granted its optimal post-hoc grouping. GPT-5, in contrast, often under-segments and at times leaves the scene undecomposed altogether, so the pipeline vectorizes the input as a whole without amodal recovery, which explains its high whole-image fidelity despite poor grouping. Qwen's lower performance mainly stems from its limited box localization during decomposition, and this spatial inaccuracy cascades directly into functional editability. Such differences arise because the backbone must localize sub-parts before our pipeline converts them into editable groups, so structural quality naturally follows its spatial reasoning capability. This confirms that while our top-down strategy inherently improves structural integrity, a spatially precise agent such as Gemini-3-flash is essential for maximizing semantic atomicity.

\begin{figure}[t]
    \centering
    \includegraphics[width=1.0\linewidth]{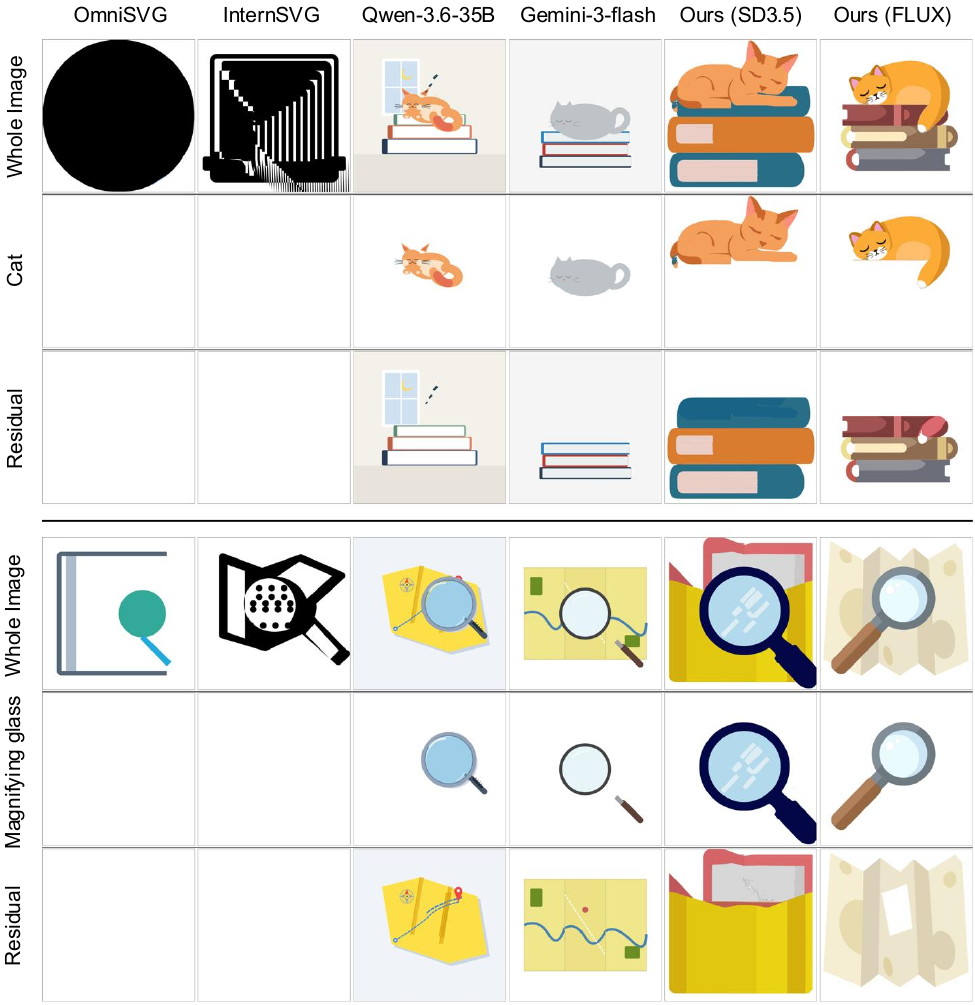}
    \caption{\textbf{Qualitative Results on Text-to-SVG task.} Generated SVGs for the prompts ``\textit{A cat sleeping on top of a stack of books}'' (top) and ``\textit{A magnifying glass over a folded map}'' (bottom). For each scenario, we show the whole image, the isolated semantic object, and the residual background. OmniSVG and InternSVG produce no semantic grouping, so their object and residual rows are empty.}
    \label{fig:t2svg_result}
\end{figure}

\noindent\textbf{Impact of Generative Occlusion Recovery}
\begin{table*}[t]
\centering
\caption{\textbf{Ablation on the Inpainting Module.} We evaluate the impact of utilizing amodal inpainting during the semantic grouping process. Grouping quality metrics for our method are measured natively on explicitly \textit{predicted} grouping structures.}
\label{tab:inpainting_ablation}
\scalebox{0.8}{%
\begin{tabular}{cccccccccc}
\toprule
\multirow[b]{2}{*}{Inpainting} & \multicolumn{3}{c}{Grouping quality (MSE)$\downarrow$} & \multicolumn{3}{c}{Grouping quality (DINO)$\uparrow$} & Editability & \multicolumn{2}{c}{Whole Image} \\
\cmidrule(lr){2-4} \cmidrule(lr){5-7} \cmidrule(lr){8-8} \cmidrule(lr){9-10}
& Recall & Precision & F1 & Recall & Precision & F1 & PERE$\downarrow$ & MSE$\downarrow$ & DINO$\uparrow$ \\
\midrule

\xmark            & .0269 & .0258 & .0263 & .9337 & .9282 & .9310 & .0307 & .0063 & .9844 \\
\cmark~  & \textbf{.0174} & \textbf{.0195} & \textbf{.0155} & \textbf{.9487} & \textbf{.9426} & \textbf{.9453} & \textbf{.0210} & \textbf{.0049} & \textbf{.9845} \\

\bottomrule
\end{tabular}%
}
\end{table*}
We evaluate the necessity of our prompt-driven amodal inpainting module by comparing our pipeline with and without it in \cref{tab:inpainting_ablation}.
When the inpainting module is disabled (\xmark), the pipeline still predicts semantic groups but fundamentally acts as a standard mask-and-crop system, strictly extracting only visible pixels.
As a result, both grouping quality and functional editability deteriorate.
Enabling generative occlusion recovery (\cmark) improves every metric, most notably reducing Grouping F1 in MSE by 41.1\% and PERE by 31.6\%.
This demonstrates that recovering the full, unoccluded geometry of a semantic component ensures the generated SVG paths accurately align with the true physical structure of the ground-truth objects, rather than just their fragmented, visible remnants.

\noindent\textbf{Evaluation on Text-to-SVG Generation}\ \ \
\begin{table}[t]
\centering
\caption{\textbf{Quantitative evaluation on Text-to-SVG task.} Evaluated on 300 prompts from MMSVG-Bench following its protocol. Ours (SD3.5) and Ours (FLUX) denote our pipeline using SD3.5-medium and FLUX.1-dev as the text-to-image model, respectively.}
\label{tab:text_to_svg_evaluation}
\resizebox{\columnwidth}{!}{%
\begin{tabular}{lcccc}
\toprule
Methods & FID$\downarrow$ & CLIP$\uparrow$ & HPSv2$\uparrow$ & Aesthetic$\uparrow$ \\
\midrule
OmniSVG            & \textbf{138.91} & 0.2435 & 0.2328 & 4.558 \\
InternSVG          & 146.18 & 0.2530 & 0.2343 & 4.592 \\
Qwen3.6-35B-A3B    & 184.72 & 0.2900 & 0.2611 & 5.075 \\
Gemini-3-flash     & 166.03 & 0.3060 & 0.2673 & 5.016 \\
\midrule
Ours (SD3.5) & 157.91 & 0.3017 & 0.2657 & 5.051 \\
Ours (FLUX)  & 171.72 & \textbf{0.3076} & \textbf{0.2732} & \textbf{5.248} \\
\bottomrule
\end{tabular}%
}
\end{table}
\cref{tab:text_to_svg_evaluation} presents the Text-to-SVG quantitative evaluation. Both of our variants substantially outperform the baselines, with FLUX achieving the best CLIP, HPSv2, and Aesthetic scores. While OmniSVG reports the lowest FID, this is an evaluation artifact: the ground-truth images used to compute FID are derived directly from its own training data, biasing the metric toward that distribution.
Visual comparisons in \cref{fig:t2svg_result} further support these results. OmniSVG and InternSVG fail entirely, and although Qwen and Gemini produce reasonable layouts, individual objects are often rendered with low fidelity. Both of our variants instead synthesize the scene faithfully. Furthermore, separating a generated SVG into its predicted foreground and residual background shows that the occluded geometry behind the foreground remains structurally complete, thanks to our prompt-driven amodal inpainting. These results show that our framework extends naturally to Text-to-SVG, producing outputs that are faithful to the prompt while retaining the same semantic grouping and geometric completeness.  

\section{Conclusion}
In this work, we address a fundamental limitation of existing SVG generation methods, which produce flat, semantically agnostic collections of paths that are difficult to manipulate. To resolve this, we reframe the task as semantic compositional SVG generation, where a multi-role VLM recursively decomposes a scene into geometrically complete, language-aligned components. To evaluate this new paradigm, we introduce the human-annotated Semantic SVG Benchmark together with the sub-component metrics Semantic Recall/Precision and Post-Edit Rendering Error. Our pipeline natively generates semantic structures that surpass the optimal post-hoc groupings of existing baselines in both grouping quality and functional editability, while preserving state-of-the-art visual fidelity. We believe this establishes a foundation for producing practically manipulable vector graphics.

\paragraph{Limitations}
The main limitation of our framework is computational cost.  On our 203-image curated benchmark, each image required on average 16.3 Gemini-3-flash calls (3.5 decompose ${+}$ 3.9 occlusion judge ${+}$ 7.1 polygon ${+}$ 1.8 pick), costing \$0.21 per image (median \$0.20; total \$42.06).  End-to-end latency averaged 255\,s per image (median 254\,s), of which 94.9\% was spent on sequential FLUX.1-Fill-dev inference (${\sim}15$\,GB bf16 GPU, 30 denoising steps at $1024{\times}1024$, four polygon variants per occluded part) and only 5.0\% on hierarchical decomposition.  Since our pipeline is inherently modular, this bottleneck can be largely mitigated by asynchronous processing across stages.

\paragraph{Acknowledgements}
This research was supported by MSIT (IITP: IITP-2026-RS-2024-00436857, RS-2024-00398115, IITP-2026-RS-2020-II201819, IITP-2026-RS-2025-02304828, RS-2026-25507282, RS-2026-25585074; NRF: RS-2025-23523979), MCST (KOCCA: RS-2026-25506607, RS-2024-00345025), MND (RS-2026-25551943) and MSS (RS-2026-25549378).
This research was also supported by the “Advanced GPU Utilization Support Program” funded by the Government of the Republic of Korea (MSIT).

\bibliography{custom}

\clearpage
\appendix

\begin{table*}[t]
\centering
\small
\caption{\textbf{Comparison with Existing SVG Benchmarks.}
We compare the domains, availability of semantic group annotations, and evaluation metrics of existing SVG benchmarks.}
\label{tab:benchmark_comparison}

\resizebox{\linewidth}{!}{%
\begin{tabular}{llcl}
\toprule

\makecell[c]{Dataset}
& \makecell[c]{Domains}
& \makecell[c]{Group\\Annotations}
& \makecell[c]{Metrics} \\

\midrule
\textbf{Ours}
& icon / illustration / emoji
& \cmark
& MSE, DINO, \textbf{Grouping quality}, \textbf{PERE} \\

SVG-Bench
& icon / font / emoji / diagram
& \xmark
& MSE, SSIM, LPIPS, DINO \\

MMSVG-Bench
& icon / illustration / character
& \xmark
& MSE, SSIM, LPIPS, DINO \\

SArena
& icon / illustration / chem / anime
& \xmark
& CLIP-I2I, SSIM, LPIPS, DINO \\

VectorGym
& icon / font / emoji / diagram
& \xmark
& MSE, LPIPS, DINO, VLM-Judge \\

\bottomrule
\end{tabular}%
}
\end{table*}

\section{Semantic SVG Benchmark Details}
\label{app:benchmark}

\subsection{Comparison with Existing SVG Datasets}
\label{app:dataset-comparison}

\cref{tab:benchmark_comparison} compares our benchmark with
SVG-Bench~\cite{starvector}, MMSVG-Bench~\cite{omnisvg},
SArena~\cite{internsvg}, and VectorGym~\cite{vectorgym} in terms of
their domains, semantic group annotations, and evaluation metrics.
These benchmarks cover diverse SVG domains but do not provide
semantic group annotations. Our benchmark includes path-level
semantic group annotations and supports component-level grouping
metrics and PERE in addition to whole-image MSE and DINO.

\subsection{Source Data and Licenses}
\label{app:data-sources}

\begin{table}[t]
\centering
\small
\caption{\textbf{Source data and licenses.} Immediate collection points for the 203 SVGs in our benchmark, with the number of samples and license of each source.}
\label{tab:data_sources}

\resizebox{\columnwidth}{!}{%
\begin{tabular}{llcl}
\toprule
Stream & Source & \# & License \\
\midrule
\multirow{2}{*}{SArena}
& SArena-Icon & 41 & \multirow{2}{*}{research use} \\
& SArena-Illustration & 23 & \\

\midrule
SVG-Bench & SVG-Emoji (test) & 29 & research use \\
\midrule
\multirow{7}{*}{Web-curated}
 & Openclipart  & 30 & CC0-1.0 \\
 & Twemoji      & 22 & CC-BY-4.0 \\
 & Noto Emoji   & 26 & Apache-2.0 \\
 & EmojiTwo     & 19 & CC-BY-4.0 \\
 & Fluent Emoji &  9 & MIT \\
 & OpenMoji     &  4 & CC-BY-SA-4.0 \\

\midrule
\multicolumn{2}{l}{\textbf{Total}} & \textbf{203} & \\
\bottomrule
\end{tabular}%
}
\end{table}

We surveyed the SVG-based evaluation sets of prior SVG generation
work, including SArena~\cite{internsvg},
SVG-Bench~\cite{starvector}, MMSVG-Bench~\cite{omnisvg}, and the test
set of LayerPeeler~\cite{layerpeeler}. After cross-set
de-duplication and the filtering described in
\cref{app:construction}, the retained samples from prior benchmarks
came from only two sources: the icon and illustration subsets of
SArena and the SVG-Emoji test subset of StarVector's SVG-Bench. These
samples constitute the re-annotated portion of our benchmark.

The remaining samples were collected directly from the primary
repositories of publicly available emoji and clipart projects:
Openclipart\footnote{\url{https://openclipart.org}},
Twemoji\footnote{\url{https://github.com/jdecked/twemoji}},
Noto Emoji\footnote{\url{https://github.com/googlefonts/noto-emoji}},
EmojiTwo\footnote{\url{https://github.com/EmojiTwo/emojitwo}},
Fluent Emoji\footnote{\url{https://github.com/microsoft/fluentui-emoji}},
and OpenMoji\footnote{\url{https://github.com/hfg-gmuend/openmoji}}.
None of these web-curated samples are byte-identical to an SVG in the
surveyed benchmark corpora.

\cref{tab:data_sources} reports the number of samples and license for
each source. All sources permit research use, and each license was
verified individually. OpenMoji is distributed under
CC-BY-SA-4.0 and therefore requires attribution upon redistribution.
The semantic decomposition and occlusion/amodal annotations are
created by us and are not present in any of the source datasets.

\subsection{Benchmark Construction Process}
\label{app:construction}

\paragraph{Sample filtering}
Following the criteria described in the main text, we inspect each
candidate's rendering and path organization. This visual inspection is
necessary because object-level partitionability cannot be determined
from SVG markup alone: in particular, a single path may span multiple
semantic objects. We remove such samples, as well as indivisible compositions, 
compositions without inter-object occlusion, 
and optimization-based vectorizations whose paths
follow color regions rather than object boundaries.

\paragraph{Annotators}
The annotation was carried out by two graduate students with an engineering background, who were compensated by our laboratory for the hours spent on the task.

\paragraph{Annotation interface}
\begin{figure}[t]
  \centering
  \includegraphics[
    width=\linewidth,
    height=0.70\textheight,
    keepaspectratio
  ]{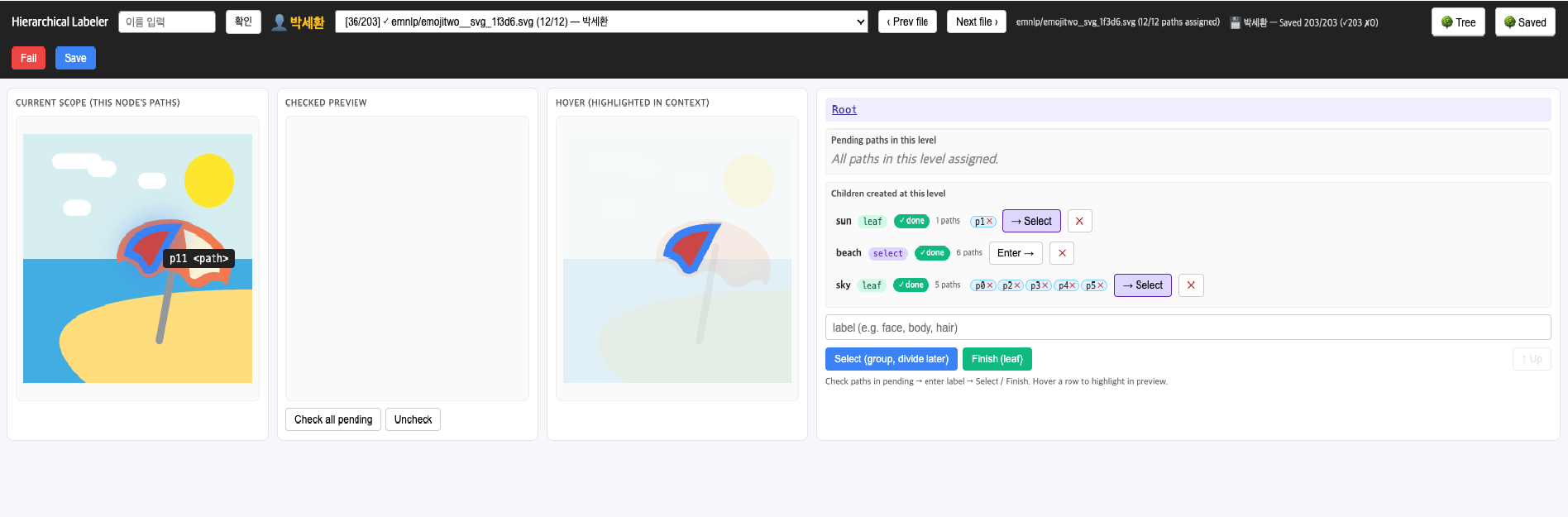}
  \caption{\textbf{Interface for Hierarchical Semantic Group Annotation.}
  Annotators inspect the rendered SVG, highlight individual paths,
  and assign them to nodes in a semantic decomposition tree.}
  \label{fig:annotation_web_ui}
\end{figure}
A custom web interface (\cref{fig:annotation_web_ui}) displays the
rendered image alongside its path list. Hovering over a path highlights
its corresponding region, and clicking assigns it to a semantic group.
Annotators use this interface to perform the recursive decomposition
described in the main text. Each leaf record stores a free-text label
(e.g., \textit{sun}, \textit{beach}, or \textit{sky}) and the exact
set of constituent paths. For an occluded object, paths are assigned
according to the identity of the object they define rather than to the
foreground occluder, preserving amodal grouping.

\paragraph{Annotation guideline}
Annotators were instructed to decompose each scene in a strict top-down manner. Starting from the whole image, they divided it into its largest semantically meaningful components, and for each resulting node recursively decided whether a further meaningful split was possible, marking the node as a leaf when no such split remained. The annotation was complete only when every branch terminated in a leaf.

A node was split only when each resulting part carries an independent semantic identity, judged by viewing the part both in isolation and in the context of the full image. A region qualifies when it can be named as a concept in the scene even if its shape alone is ambiguous; a plain blue rectangle, for instance, is recognized as the sea once a surfer is shown riding on it. Conversely, a part is kept within its parent when it is meaningful only in combination with the object it belongs to, as with the limbs of a person or the strap of a hat. Under the same criterion, an object such as a hat is separated from the person wearing it, since it retains its identity on its own. Elements without independent semantic meaning, such as shadows and outlines, were either absorbed into the group they belong to or labeled \textit{none}; nodes labeled \textit{none} are excluded from evaluation. Each group was assigned a free-text label describing the entity it represents. Labels were not drawn from a fixed vocabulary, since the benchmark spans diverse visual domains. The full guideline document is released together with the benchmark.

\paragraph{Validation and release}
After annotation, we manually inspect each completed semantic tree and its primitive assignments. We retain a sample only when every drawable primitive is assigned to exactly one leaf, such that the leaf primitive sets are pairwise disjoint and jointly cover the complete set of drawable primitives. This constraint concerns primitive ownership; the rendered regions of different leaves may still overlap spatially because of occlusion. Candidates for which no such assignment can be constructed are excluded. In total, 203 SVGs pass filtering, annotation, and manual validation. Each released sample contains the raw SVG, its decomposition tree, provenance information, and license.

\section{Implementation Details}
\label{app:implementation-details}

\subsection{Details on the Text-to-SVG Extension}
\label{app:train_details}

\paragraph{Text-to-image model training.}
For the Text-to-SVG task, we adapt two general-purpose text-to-image models, FLUX.1-dev~\cite{fluxdev} and SD3.5-medium~\cite{sd3.5}, to icon- and illustration-style raster generation, and feed the generated raster into our pipeline. Both models are trained on 32{,}000 images sampled from MMSVG-Icon and MMSVG-Illustration~\cite{omnisvg}, rasterized at $1024\times1024$ on a white background. Each caption is the dataset description prefixed with a trigger token \texttt{<vector>}, which is also prepended at inference.

We attach rank-128 LoRA adapters~\cite{lora} to the attention and feed-forward projections of every transformer block, keeping the text encoders, VAE, and all remaining layers frozen; this yields 523.0M trainable parameters for FLUX.1-dev and 188.0M for SD3.5-medium. Both models are trained at $1024\times1024$ with AdamW at a learning rate of $10^{-4}$ and an effective batch size of 64. Since FLUX.1-dev is guidance-distilled, it is trained without classifier-free guidance~\cite{cfg}, whereas SD3.5-medium drops the caption with probability 0.1 so that its unconditional branch is adapted as well.

\paragraph{Checkpoint selection and inference.}
We use the checkpoint after one epoch for FLUX.1-dev and two epochs for SD3.5-medium. At inference, images are generated at $1024\times1024$ using the default settings of each model card: 28 sampling steps with guidance scale 3.5 for FLUX.1-dev, and 40 steps with CFG scale 4.5 for SD3.5-medium.

\subsection{Evaluation Details}
\label{app:eval_details}

\paragraph{Common Representation and Rendering.}
Following \cref{sec:eval_grouping}, $\mathcal{G}$ contains all labeled
non-root nodes obtained by flattening the GT semantic tree; nodes
labeled \texttt{none} are excluded. Each $g_i\in\mathcal{G}$ is
rendered from the drawable primitives in its subtree. Drawable
elements are indexed in SVG document order over \texttt{path},
\texttt{rect}, \texttt{circle}, \texttt{ellipse}, \texttt{line},
\texttt{polyline}, and \texttt{polygon}; elements inside non-rendering
definition containers are excluded.

All component images are rasterized with CairoSVG~2.9.0 at
$256{\times}256$, alpha-composited onto an opaque white background,
and scaled to $[0,1]$. Component-level scores are computed on a crop
derived from the union bounding box of the non-white pixels in the two
renderings. MSE is averaged over all pixels and RGB
channels. For DINO, we use mean-pooled tokens from the DINOv2-base model
\citep{oquab2023dinov2} and report the rescaled cosine similarity
$(1+\cos)/2$.

\paragraph{Greedy Approximation of Optimal Grouping.}
We approximate the subset optimization in
\cref{eq:optimal_group} using a two-phase greedy search. Exact
enumeration requires evaluating $2^K$ subsets and is intractable for
SVGs containing hundreds or thousands of generated paths. We denote
the subset returned for $g_i$ by $\tilde{p}_i$, distinguishing it from
the ideal optimizer $p_i^\star$ in \cref{eq:optimal_group}.

For each $g_i$, we retain as candidates only generated paths whose
isolated non-white mask has nonzero overlap with
$\mathcal{R}(g_i)$ at $256{\times}256$, implemented as mask IoU
$>10^{-9}$. Starting from the empty subset, we traverse these
candidates in SVG document order and retain a path if adding it
strictly improves the current objective. We then perform at most ten
rounds of alternating refinement. In each round, all unselected
candidates are considered for addition, followed by all selected
candidates for removal; a change is accepted only if it strictly
improves the objective. The search terminates when a complete round
leaves the subset unchanged.

After selecting $\tilde{p}_i$ using MSE, we evaluate the same subset
with both MSE and DINO similarity. Thus, DINO provides a complementary
semantic similarity score for the MSE-selected grouping rather than
inducing a separate path search. PERE likewise uses this MSE-selected
subset.

The term \textit{optimal grouping} refers to the combinatorial target
defined in \cref{eq:optimal_group}; the reported results are greedy
approximations rather than certified global optima. Moreover, the
search is performed independently for each $g_i$. The resulting
subsets $\tilde{p}_i$ may therefore overlap and are not constrained to
form a global partition of the generated SVG.

\paragraph{Predicted Grouping and Bidirectional Matching.}
In the \textit{predicted} setting, we flatten the generated semantic
hierarchy into the candidate groups $\mathcal{P}$ defined in
\cref{sec:eval_grouping}, retaining non-root groups at every depth.
For each metric, we construct the complete
$|\mathcal{G}|\times|\mathcal{P}|$ score matrix. Semantic Recall
selects the best predicted group in each GT row, whereas Semantic
Precision selects the best GT node in each predicted-group column.
These are independent nearest-neighbor matches rather than a
one-to-one assignment; hence, the same predicted group may match
multiple GT nodes, and the same GT node may match multiple predicted
groups. MSE and DINO construct and match their score matrices
independently.

Recall, Precision, and F1 are first computed within each image. F1 is
the harmonic mean of that image's Recall and Precision. The MSE
version is applied to the two errors and remains lower-is-better,
whereas the DINO version is applied to the two similarities and
remains higher-is-better. Dataset-level results are then
macro-averaged over the 203 images. Consequently, the reported F1 is
not generally equal to the harmonic mean of the dataset-level Recall
and Precision values.

\paragraph{Post-Edit Rendering Error.}
For PERE, we construct occlusion cases from pairs of depth-1 GT
groups defined in \cref{sec:eval_editability}, i.e., the root's
immediate children, whose amodal masks overlap. To determine the
foreground object, we delete each group from the GT SVG in turn and
measure the mean pixel change within the overlapping region relative
to the complete GT rendering. The group whose deletion produces the
larger change is treated as the foreground occluder $o$; ties are
resolved by GT node order.

Operationally, $\mathcal{O}$ is treated as a collection containing one
foreground occluder for each overlapping pair. Thus, the same occluder
may occur multiple times when it overlaps multiple objects. In the
\textit{optimal} setting, $\hat{o}$ is the MSE-selected subset
$\tilde{p}_i$ constructed for the corresponding GT occluder $o$. 
In the \textit{predicted} setting, for PERE we restrict
$\mathcal{P}$ to the outermost labeled groups in the generated SVG,
and $\hat{o}$ is the group with the lowest full-frame MSE to $o$, as
defined in \cref{eq:pere_match}.

For each case, we remove the paths of $o$ from the GT SVG and those of
$\hat{o}$ from the generated SVG. The two edited SVGs are rendered
independently at $256{\times}256$ on white, and their full-frame MSE
is computed without bounding-box cropping. Every deletion starts from
the original, unedited SVG rather than accumulating previous
deletions. PERE is averaged first over the occlusion pairs within each
image and then over the benchmark images. All 203 images contain at
least one valid pair, yielding 337 pairs in total.

\paragraph{Relation to Whole-Image Fidelity.}
Whole-image MSE and DINO are reported separately to measure whether a
method generated the target image in the first place. Optimal grouping
removes the requirement for native grouping metadata but remains
sensitive to missing or incorrectly generated primitives. Likewise,
PERE is not normalized by the unedited whole-image error and therefore
measures end-to-end editability rather than grouping alone. For
methods reported in both the \textit{optimal} and \textit{predicted}
settings, the underlying SVG is identical. 
Differences in Semantic Recall and PERE therefore reflect
the effect of using the model's predicted groups instead of
GT-conditioned path optimization.

\begin{table}[t]
\centering
\caption{\textbf{Quantitative results on the Image-to-SVG task with MMSVG-Bench.} Evaluated on 300 samples with whole-image fidelity metrics, since MMSVG-Bench provides no ground-truth semantic-group annotations. The best results are highlighted in \textbf{bold}.}
\label{tab:mmsvg}
\resizebox{\linewidth}{!}{%
\begin{tabular}{lcccc}
\toprule
& \multicolumn{4}{c}{Whole Image} \\
\cmidrule(lr){2-5}
Methods & MSE$\downarrow$ & LPIPS$\downarrow$ & SSIM$\uparrow$ & DINO$\uparrow$ \\
\midrule
VTracer          & .0021 & .0357 & .9650 & .9934 \\
OmniSVG          & .0546 & .2031 & .8783 & .9202 \\
InternSVG        & .0254 & .1114 & .8664 & .9646 \\
StarVector       & .0592 & .2502 & .8857 & .8604 \\
LayerPeeler      & .0660 & .2212 & .8345 & .7846 \\
Qwen3.6-35B-A3B  & .0797 & .3419 & .8479 & .9118 \\
Gemini-3-flash & .0409 & .1961 & .8907 & .9559 \\
\midrule
Ours             & \textbf{.0016} & \textbf{.0278} & \textbf{.9701} & \textbf{.9958} \\
\bottomrule
\end{tabular}%
}
\end{table}

\section{Evaluation on Image-to-SVG with MMSVG-Bench} \label{app:mmsvg_bench}
To further validate that our performance is not confined to our curated benchmark, we evaluate the Image-to-SVG task on MMSVG-Bench~\cite{omnisvg} against all baselines. Since MMSVG-Bench provides no ground-truth semantic-group annotations, grouping quality and editability cannot be measured on it; we therefore report whole-image fidelity only. As shown in \cref{tab:mmsvg}, our method achieves the best scores across all four metrics. Notably, although our pipeline decomposes the scene and reassembles it from per-component vectorizations, it still attains the highest fidelity, confirming that our strong performance extends to a standard benchmark beyond our own.

\section{More Qualitative Results}
\label{app:more_qualitative_results}
This section provides additional qualitative results complementing the analyses in the main paper. All figures follow the same format as their counterparts in the main paper, showing more samples across diverse scenes and prompts. Across various scenarios, we consistently visualize the full generated image, the isolated semantic object, and the residual background to assess functional editability.

\begin{figure*}[p]
    \centering
    \includegraphics[
        width=\textwidth,
        height=\textheight,
        keepaspectratio
    ]{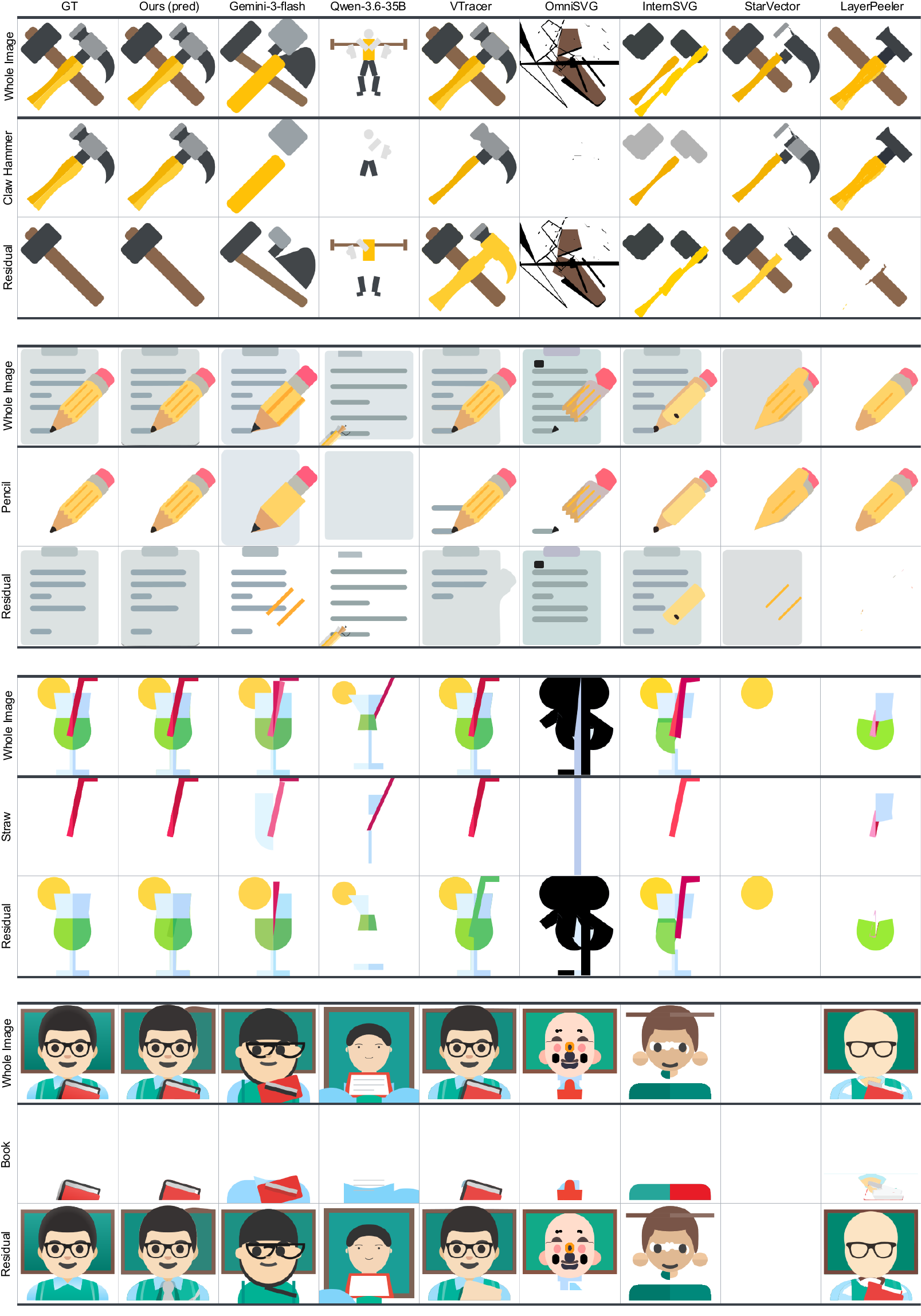}
  \caption{\textbf{Additional Qualitative Results on the Image-to-SVG Task (1/3).} For each scenario, we show the whole image, isolated semantic object, and residual background. The whole image in the GT column is also used as the input image for all methods. Ours (pred) denotes the group (set of primitives) predicted by our model. For all remaining generated outputs, we show the semantic object rendered from paths selected using an optimal post-hoc grouping strategy. Our method effectively isolates discrete entities and recovers occluded background geometry through high-quality amodal inpainting.
  }
    \label{fig:appendix_qualitative_1}
\end{figure*}

\begin{figure*}[p]
    \centering
    \includegraphics[
        width=\textwidth,
        height=\textheight,
        keepaspectratio
    ]{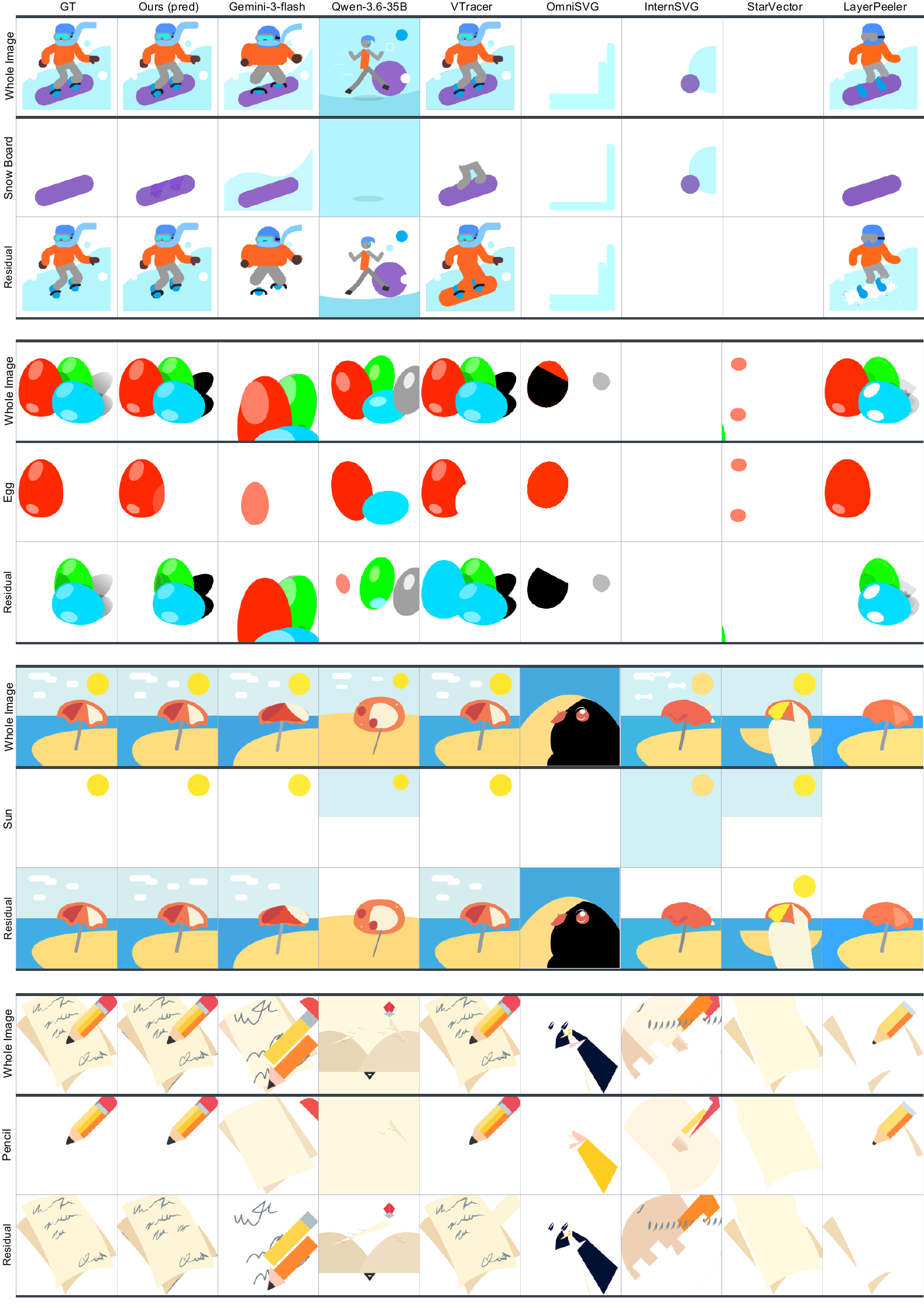}
  \caption{\textbf{Additional Qualitative Results on the Image-to-SVG Task (2/3).} For each scenario, we show the whole image, isolated semantic object, and residual background. The whole image in the GT column is also used as the input image for all methods. Ours (pred) denotes the group (set of primitives) predicted by our model. For all remaining generated outputs, we show the semantic object rendered from paths selected using an optimal post-hoc grouping strategy. Our method effectively isolates discrete entities and recovers occluded background geometry through high-quality amodal inpainting.}
    \label{fig:appendix_qualitative_2}
\end{figure*}

\begin{figure*}[p]
    \centering
    \includegraphics[
        width=\textwidth,
        height=\textheight,
        keepaspectratio
    ]{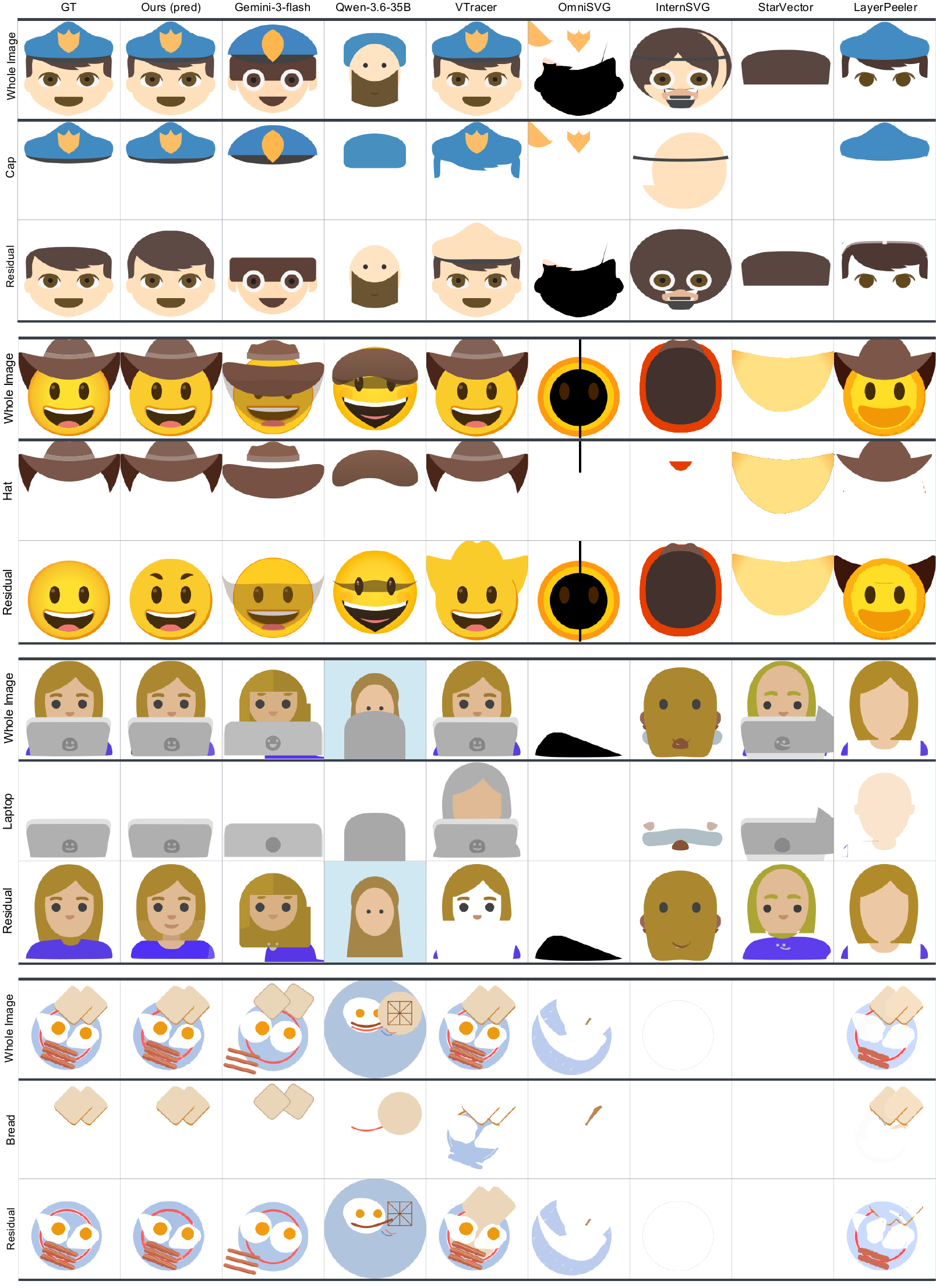}
  \caption{\textbf{Additional Qualitative Results on the Image-to-SVG Task (3/3).} For each scenario, we show the whole image, isolated semantic object, and residual background. The whole image in the GT column is also used as the input image for all methods. Ours (pred) denotes the group (set of primitives) predicted by our model. For all remaining generated outputs, we show the semantic object rendered from paths selected using an optimal post-hoc grouping strategy. Our method effectively isolates discrete entities and recovers occluded background geometry through high-quality amodal inpainting.}
    \label{fig:appendix_qualitative_3}
\end{figure*}

\begin{figure*}[p]
    \centering
    \includegraphics[
        width=\textwidth,
        height=0.9\textheight,
        keepaspectratio
    ]{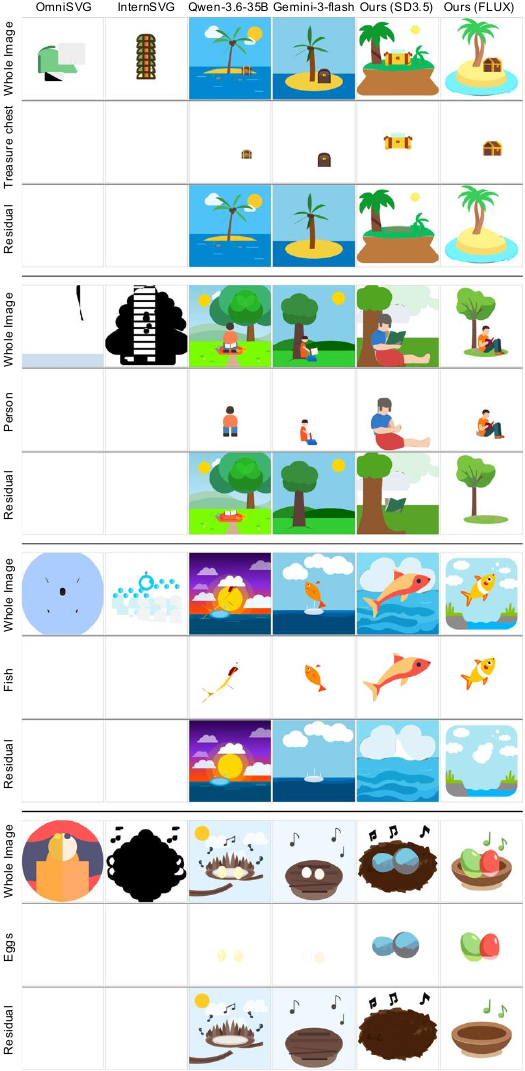}
    \caption{\textbf{Additional Qualitative Results on the Text-to-SVG Task.} Generated SVGs for the prompts ``\textit{A small island with a palm tree and treasure chest}'', ``\textit{A person reading a book under a tree}'', ``\textit{A fish jumping from water into clouds}'', and ``\textit{A bird's nest with two eggs and musical notes}'' (top to bottom). For each scenario, we show the whole image, the isolated semantic object, and the residual background. OmniSVG and InternSVG produce no semantic grouping, so their object and residual rows are empty.}
    \label{fig:appendix_qualitative_t2svg}
\end{figure*}

\cref{fig:appendix_qualitative_1,fig:appendix_qualitative_2,fig:appendix_qualitative_3} extend the Image-to-SVG comparison in \cref{fig:qualitative_result}, and \cref{fig:appendix_qualitative_t2svg} extends the Text-to-SVG comparison in \cref{fig:t2svg_result}. These additional results collectively substantiate our pipeline's structural superiority: it robustly isolates distinct entities based on semantic logic and seamlessly recovers obscured background geometries via amodal inpainting, ensuring high-quality structural completeness across diverse visual domains.

\clearpage\onecolumn\raggedbottom\section{Prompts for our tool-augmented framework}
\label{app:prompts}
\subsection{Region Decomposition Decision Prompt}
See Figure~\ref{fig:prompt_region_decomposition} for an example prompt.

\subsection{Residual Judge Prompt}
See Figure~\ref{fig:prompt_residual_judge} for an example prompt.

\subsection{Inpainting Judge Prompt}
See Figure~\ref{fig:prompt_inpainting_judge} for an example prompt.

\subsection{Polygon Prediction Prompt}
See Figure~\ref{fig:prompt_polygon_p0_baseline} for an example prompt.

\subsection{Polygon Prediction Prompt (Category-Aware)}
See Figure~\ref{fig:prompt_polygon_p1_category} for an example prompt.

\subsection{Polygon Prediction Prompt (Structured Reasoning)}
See Figure~\ref{fig:prompt_polygon_p2_structured} for an example prompt.

\subsection{Polygon Prediction Prompt (Geometric Strict)}
See Figure~\ref{fig:prompt_polygon_p3_geometric} for an example prompt.

\subsection{Inpainting Selection Prompt}
See Figure~\ref{fig:prompt_inpainting_selection} for an example prompt.

\subsection{Image-to-SVG Prompt for the VLM Baselines}
See Figure~\ref{fig:prompt_grouping_svg} for an example prompt.

\subsection{Text-to-SVG Prompt for the VLM Baselines}
See Figure~\ref{fig:prompt_text2svg} for an example prompt.

\clearpage
\raggedbottom 
\vspace{1em} 
\begin{tcolorbox}[
    breakable, 
    colframe=gray!50!black,
    colback=gray!10!white,
    title=Region Decomposition Decision Prompt,
    width=\textwidth, sharp corners=southwest,
    left=1mm, right=1mm, top=0mm, bottom=0mm
]
\scriptsize
\begin{lstlisting}[
    aboveskip=0pt,
    belowskip=0pt,
    breakatwhitespace=false,
    basicstyle=\ttfamily\scriptsize,
    showstringspaces=false,
    frame=none
]
You decide whether to decompose a region of a flat-color illustration
into named sub-parts. {parent_context}

======================================================================
DECIDE: is_atomic = true | false

ATOMIC IS THE DEFAULT. The pipeline already paints the parent's mask;
only split when each child would be drawn on its OWN LAYER by a
human designer. Visual fragments sharing one fill are NOT separate
layers.

Choose ATOMIC unless ALL THREE hold:
  1. The region contains 2+ semantically NAMED sub-parts (each can
     carry a plain noun-phrase a designer would put on a layer).
  2. EACH sub-part occupies >= 25% of the parent region's bbox area.
     This is a HARD floor. Tiny decorations (logos, buttons,
     stickers, individual pedals/seats/handles on a vehicle,
     small accessories) are NOT separate parts -- they belong to
     the parent's fill. If any candidate child would be < 25%
     area, the whole split is wrong -> ATOMIC.
  3. The sub-parts visibly OCCLUDE each other (one silhouette has a
     notch cut by the other, so the inpaint merge needs separate
     layers to reconstruct each).

DEPTH-2 GATE -- depth-2 is RARE.

When parent_label is non-empty you are at depth 2. ATOMIC is almost
always the right answer here -- the dataset has roughly 1 depth-2
case in 15. Only mark non-atomic if EVERY check above holds AND the
split is obvious to ANY designer reading the parent shape. Any
uncertainty -> ATOMIC.

Depth >= 3 is BLOCKED by code -- never attempt.

======================================================================
ALWAYS ATOMIC (no exceptions, at any depth):
  - Single-fill natural shapes (sun, moon, cloud, fire, smoke, water,
    leaf, flower, rainbow band, individual star, single mushroom,
    single mountain, single tree).
  - Tools / man-made objects (laptop, phone, hammer, bottle, lamp,
    trophy, spray bottle, magnifying glass, target, lightbulb,
    sunglasses, kettle, microphone, vase, basket, knife, bicycle).
    Decorations on their surface (logo, label, sticker, screen
    content, key rows, individual pedals/seats/handles) are part
    of the tool's fill -- NEVER separate parts.
  - ALL body parts including HEAD (head, face, hand, foot, eye, ear,
    nose, mouth, hair, lip, neck). NEVER split a head into face
    features. Characters / persons / animals are ATOMIC as a single
    body -- do not split into head/body/legs.
  - Repeated identical small shapes (spikes, gems, dots, stripes,
    petals, rays, sparkles, buttons). Keep as one plural entry OR
    omit entirely. Do not enumerate per instance.
  - Pure stroke / outline / shading / highlight / shadow. Never split.

NEVER EMIT placeholder labels: "background", "none", "remainder",
"residual", "rest", "leftover", "other", "misc", "outline", "stroke",
"highlight", "shadow", "shading", "fill". The parent mask absorbs
unnamed pixels -- do not invent catch-alls.

======================================================================
Z-ORDER -- HARD RULE (inpaint merge depends on this).

List `components` BACK-to-FRONT (painter's order; index 0 deepest):
  - Background / container goes FIRST. Foreground / overlay LAST.
  - Occlusion decides: if A is partially hidden by B, A comes
    before B. Whichever silhouette has a notch cut into it is BEHIND.
  - Frame / outline around a fill goes FIRST.
  - Container + contents: the container goes FIRST, contents LAST.
  - Repeated instances side-by-side without overlap -- natural reading
    order (left -> right) is fine.

======================================================================
LABEL + BBOX RULES:
  - label = plain noun phrase. No parent tokens, no catch-all.
  - _bbox = tight [ymin, xmin, ymax, xmax] in 0-1000 normalized
    coords of the FULL ORIGINAL image.
  - Sibling bboxes describe DISJOINT regions (may touch, must not
    strongly overlap). EXCEPT container + contents at the same level
    (jail + prisoner, sunglasses + face) -- overlap OK, z-order
    disambiguates.

======================================================================
OUTPUT -- valid JSON only, no markdown:
{{
  "is_atomic": <true|false>,
  "components": [{{"label": "...", "_bbox": [ymin, xmin, ymax, xmax]}}, ...]
}}

======================================================================
EXAMPLES

[Root, scene with sun + two clouds (multi-entity, all atomic).]
{{
  "is_atomic": false,
  "components": [
    {{"label": "sun",         "_bbox": [120, 700, 320, 920]}},
    {{"label": "left cloud",  "_bbox": [400,  80, 600, 420]}},
    {{"label": "right cloud", "_bbox": [380, 540, 580, 880]}}
  ]
}}

[Atomic, focus on "laptop" -- single tool.]
{{"is_atomic": true, "components": []}}

[Atomic, focus on "head" -- face / eyes / mouth / hair are NEVER separated.]
{{"is_atomic": true, "components": []}}

[Atomic, focus on any sub-region with parent_label non-empty --
default answer at depth 2 is ATOMIC.]
{{"is_atomic": true, "components": []}}
\end{lstlisting}
\end{tcolorbox}
\captionof{figure}{\textbf{Region Decomposition Decision Prompt.}}
\label{fig:prompt_region_decomposition}
\vspace{0.5em}
\begin{figure}[H]
\begin{tcolorbox}[
    colframe=gray!50!black,
    colback=gray!10!white,
    title=Residual Judge Prompt,
    width=\textwidth, sharp corners=southwest,
    left=1mm, right=1mm, top=0mm, bottom=0mm
]
\scriptsize
\begin{lstlisting}[
    aboveskip=0pt,
    belowskip=0pt,
    breakatwhitespace=false,
    basicstyle=\ttfamily\scriptsize,
    showstringspaces=false,
    frame=none
]
You decide ONE thing about a vector-graphic layer.

INPUTS
- IMAGE 1: the full original scene (for context).
- IMAGE 2: a single layer labeled "{label}", cut out on white.
  White pixels = transparent / not part of the layer.

PIXEL CONTEXT
- Visible pixels of THIS layer in IMAGE 2: {part_visible_px}
- Total color (non-white) pixels of the FULL scene in IMAGE 1: {scene_color_px}
- This layer covers about {ratio_pct:.2f}% of the scene's color area.

DECISION -- is_residual
  Set TRUE only when IMAGE 2 is GARBAGE leftover that a designer
  would never put on a layer:
    - Scattered dots / noise pixels with no shape.
    - Thin edge slivers / partial outlines tracing the boundary of
      a sibling part.
    - Broken fragments of a single shape already represented by
      siblings.

  Set FALSE -- this layer is REAL -- when IMAGE 2 is any coherent
  shape, INCLUDING:
    - A coherent large fill with CUTOUTS where foreground siblings
      sit on top of it. This is the SCENE BACKGROUND
      (sky / circular plate / sand ground / wall). The fact that
      it has notches cut into it does NOT make it residual -- it
      makes it the back layer of the painter's stack.
    - A partially-occluded coherent sub-shape.

The phrase "leftover background partition with cutouts" describes a
REAL background layer, NOT residual. Background layers paint FIRST
and other parts paint on top of them; their characteristic look is
"big fill with foreground-shaped holes". Classify them as is_residual
= FALSE so the downstream stage-2 judge can decide z-order.

PRIORITY HINTS (apply BEFORE the qualitative judgement):
  - If the layer covers >= 15% of the scene's color area, it is
    almost certainly a real layer, not slivers. Default FALSE.
  - If the layer is mostly thin lines / scattered dots covering
    < 5% of the scene, it is almost certainly residual. Default TRUE.

Be CONSERVATIVE: when uncertain, FALSE (keep). Dropping a real layer
is irreversible; keeping a borderline layer just adds one more
sibling.

OUTPUT -- JSON only, no markdown:
{{"is_residual": true | false, "reason": "one short sentence naming the visible evidence"}}

\end{lstlisting}
\end{tcolorbox}
\caption{\textbf{Residual Judge Prompt.}}
\label{fig:prompt_residual_judge}
\end{figure}

\vspace{0.5em}
\begin{figure}[H]
\begin{tcolorbox}[
    colframe=gray!50!black,
    colback=gray!10!white,
    title=Inpainting Judge Prompt,
    width=\textwidth, sharp corners=southwest,
    left=1mm, right=1mm, top=0mm, bottom=0mm
]
\scriptsize
\begin{lstlisting}[
    aboveskip=0pt,
    belowskip=0pt,
    breakatwhitespace=false,
    basicstyle=\ttfamily\scriptsize,
    showstringspaces=false,
    frame=none
]
You decide up to FOUR things about a vector-graphic layer.

INPUTS
- IMAGE 1: the full original scene.
- IMAGE 2: a single layer labeled "{label}", cut out on white.
  White pixels = transparent / not part of the layer. IMAGE 2 may be
  incomplete because in IMAGE 1 some of this layer is hidden under
  another part drawn on top of it.

PART PIXEL CONTEXT
- Visible pixels of THIS layer in IMAGE 2: {part_visible_px}
- Total color (non-white) pixels of the FULL scene in IMAGE 1: {scene_color_px}
- This layer covers about {ratio_pct:.2f}% of the scene's color area.
{sibling_block}{parent_block}
DECISION A -- is_residual
  Set TRUE when IMAGE 2 is scatter / dot fragments / thin slivers /
  leftover pixels with no coherent shape -- i.e. there is no single
  named sub-part a designer would draw to produce this silhouette.
  Set FALSE when IMAGE 2 forms a meaningful coherent sub-shape (even
  if partially occluded).

DECISION B -- occluded (evaluate ONLY when is_residual=false)
  Set TRUE when ALL of:
    (a) IMAGE 1 shows clear visual evidence that this layer is occluded
        by another scene element -- IMAGE 2's silhouette has a noticeable
        chunk cut out by something drawn in front of it.
    (b) The hidden region is non-trivial: rough rule, adding it would
        need pixels comparable to >= 5% of the layer's currently visible
        area. Thin antialiased edges and sub-pixel noise do NOT count.
  Set FALSE otherwise (already complete / in-front of everything that
  touches it / supposed occlusion is just edge noise).

Be CONSERVATIVE on occluded: prefer false unless the gap is clearly
visible. When is_residual=true, ALWAYS set occluded=false (we drop
the residual; we don't reconstruct it).

DECISION C -- occluding_siblings (evaluate ONLY when occluded=true)
  Output a JSON array of SIBLING NAMES (the 'name' column of the table
  above) for every sibling that paints OVER this layer. Closed set: only
  those exact names are valid, NEVER invent new names and NEVER pick the
  parent layer. Use the geometric prior:
    - High `sib->tgt` with a visibly cut-out silhouette in IMAGE 1 =>
      strong occluder candidate.
    - High `iou` with symmetric coverage => probably a redundant pair,
      not an occluder.
    - `adj_px > 0` with `sib->tgt ~ 0` => they only touch, not occlude.
  Order entries most-occluding first. Empty array `[]` is required when
  occluded=false; it is also acceptable when occluded=true and you
  cannot identify the occluder from the provided list (the prompt fails
  open to FLUX in that case).
{decision_d_block}
OUTPUT -- JSON only, no markdown:
{"is_residual": true | false, "occluded": true | false, "occluding_siblings": [<sibling names>], "reason": "one short sentence naming the visible evidence"{description_field}}

\end{lstlisting}
\end{tcolorbox}
\caption{\textbf{Inpainting Judge Prompt.}}
\label{fig:prompt_inpainting_judge}
\end{figure}

\vspace{0.5em}
\begin{figure}[H]
\begin{tcolorbox}[
    colframe=gray!50!black,
    colback=gray!10!white,
    title=Polygon Prediction Prompt,
    width=\textwidth, sharp corners=southwest,
    left=1mm, right=1mm, top=0mm, bottom=0mm
]
\scriptsize
\begin{lstlisting}[
    aboveskip=0pt,
    belowskip=0pt,
    breakatwhitespace=false,
    basicstyle=\ttfamily\scriptsize,
    showstringspaces=false,
    frame=none
]
You are given two images:
  IMAGE 1: the full original scene.
  IMAGE 2: a cropped layer showing ONLY the part labeled "{label}"
           (white = transparent / not part of the layer). It is
           INCOMPLETE because other scene elements occlude it.
Coordinates: 0..1000 normalized over IMAGE 1 (origin top-left).

Infer the complete true shape of "{label}" as it exists in IMAGE 1,
reasoning ONLY from the visible evidence in IMAGE 2 (visible contour
curvature, proportions, symmetry) together with where occluders hide it
in IMAGE 1. Make NO assumption about object category or any typical /
textbook shape -- derive the shape from what is actually shown. Output
ONE closed polygon of that complete shape that:
- passes through ALL visible pixels of "{label}",
- continues the visible contour smoothly where it is smooth and with
  corners where it has corners,
- fills ONLY the genuinely occluded gaps, never exceeding a visible
  natural endpoint or entering background.
Single closed boundary, 30-50 vertices, short segments on curves.

Output ONLY a JSON object (no markdown):
{{"polygon": [[x1,y1], [x2,y2], ..., [xN,yN]]}}

\end{lstlisting}
\end{tcolorbox}
\caption{\textbf{Polygon Prediction Prompt 1.}}
\label{fig:prompt_polygon_p0_baseline}
\end{figure}

\vspace{0.5em}
\begin{figure}[H]
\begin{tcolorbox}[
    colframe=gray!50!black,
    colback=gray!10!white,
    title=Polygon Prediction Prompt (Category-Aware),
    width=\textwidth, sharp corners=southwest,
    left=1mm, right=1mm, top=0mm, bottom=0mm
]
\scriptsize
\begin{lstlisting}[
    aboveskip=0pt,
    belowskip=0pt,
    breakatwhitespace=false,
    basicstyle=\ttfamily\scriptsize,
    showstringspaces=false,
    frame=none
]
You are given two images:
  IMAGE 1: the full original scene.
  IMAGE 2: a cropped layer showing ONLY the part labeled "{label}"
           (white = transparent / not part of the layer). It is
           INCOMPLETE because other scene elements occlude it.
Coordinates: 0..1000 normalized over IMAGE 1 (origin top-left).

Predict the full silhouette of "{label}" as a typical flat vector
illustration of that object would appear in IMAGE 1. Use BOTH:
  (a) the visible part in IMAGE 2 -- its actual contour, proportions,
      orientation, and where it is interrupted by an occluder,
  (b) common knowledge of what a "{label}" looks like as a clean
      illustration -- its overall outline and proportions.
Anchor the silhouette to the visible pixels (never deviate from them).
For occluded gaps, prefer the shape most consistent with (a) AND (b).
Never extend into background or past a visible natural endpoint.
Single closed boundary, 30-50 vertices, short segments on curves.

Output ONLY a JSON object (no markdown):
{{"polygon": [[x1,y1], [x2,y2], ..., [xN,yN]]}}

\end{lstlisting}
\end{tcolorbox}
\caption{\textbf{Polygon Prediction Prompt 2  (Category-Aware).}}
\label{fig:prompt_polygon_p1_category}
\end{figure}

\vspace{0.5em}
\begin{figure}[H]
\begin{tcolorbox}[
    colframe=gray!50!black,
    colback=gray!10!white,
    title=Polygon Prediction Prompt (Structured Reasoning),
    width=\textwidth, sharp corners=southwest,
    left=1mm, right=1mm, top=0mm, bottom=0mm
]
\scriptsize
\begin{lstlisting}[
    aboveskip=0pt,
    belowskip=0pt,
    breakatwhitespace=false,
    basicstyle=\ttfamily\scriptsize,
    showstringspaces=false,
    frame=none
]
You are given two images:
  IMAGE 1: the full original scene.
  IMAGE 2: a cropped layer showing ONLY the part labeled "{label}"
           (white = transparent / not part of the layer). It is
           INCOMPLETE because other scene elements occlude it.
Coordinates: 0..1000 normalized over IMAGE 1 (origin top-left).

Predict the complete outline of "{label}" as it exists in IMAGE 1,
including the parts hidden by occluders.

Reason in TWO steps before answering:
  STEP 1 -- VISIBLE: identify the contour of "{label}" that is fully
    visible in IMAGE 2. Note its curvature, corners, symmetry, and the
    natural endpoints where the silhouette is interrupted by another
    layer drawn on top.
  STEP 2 -- OCCLUDED: locate the regions in IMAGE 1 where another
    element covers "{label}". Estimate the most plausible continuation
    of the silhouette into those regions, using contour smoothness,
    symmetry, and proportional cues from STEP 1. Do not invent shape
    beyond what the visible evidence supports.

Then output ONE closed polygon that:
- passes through ALL visible pixels of "{label}",
- continues smoothly into the occluded regions,
- never enters background pixels or extends past visible endpoints.
Single closed boundary, 30-50 vertices, short segments on curves.

Output ONLY a JSON object (no markdown):
{{"polygon": [[x1,y1], [x2,y2], ..., [xN,yN]]}}

\end{lstlisting}
\end{tcolorbox}
\caption{\textbf{Polygon Prediction Prompt 3 (Structured Reasoning).}}
\label{fig:prompt_polygon_p2_structured}
\end{figure}

\vspace{0.5em}
\begin{figure}[H]
\begin{tcolorbox}[
    colframe=gray!50!black,
    colback=gray!10!white,
    title=Polygon Prediction Prompt (Geometric Strict),
    width=\textwidth, sharp corners=southwest,
    left=1mm, right=1mm, top=0mm, bottom=0mm
]
\scriptsize
\begin{lstlisting}[
    aboveskip=0pt,
    belowskip=0pt,
    breakatwhitespace=false,
    basicstyle=\ttfamily\scriptsize,
    showstringspaces=false,
    frame=none
]
You are given two images:
  IMAGE 1: the full original scene.
  IMAGE 2: a cropped layer showing the part labeled "{label}", cut out
           on a white background. White = not part of the layer. The
           layer is INCOMPLETE because other elements occlude it in
           IMAGE 1.
Coordinates: 0..1000 normalized over IMAGE 1 (origin top-left).

Reconstruct the complete silhouette of "{label}" as a SINGLE closed
polygon. Rules:
  - The polygon MUST enclose every visible pixel of "{label}" from
    IMAGE 2 -- no visible pixel may fall outside the polygon.
  - The polygon must NOT extend into regions of IMAGE 1 that are
    clearly background (white / empty space) or that belong to layers
    other than the occluders of "{label}".
  - Closures across occluded gaps should be the simplest contour
    consistent with the visible silhouette's curvature and endpoint
    tangents.
  - Place more vertices on curved regions, fewer on straight runs.
Single closed boundary, 30-50 vertices.

Output ONLY a JSON object (no markdown):
{{"polygon": [[x1,y1], [x2,y2], ..., [xN,yN]]}}

\end{lstlisting}
\end{tcolorbox}
\caption{\textbf{Polygon Prediction Prompt 4 (Geometric Strict).}}
\label{fig:prompt_polygon_p3_geometric}
\end{figure}

\vspace{0.5em}
\begin{figure}[H]
\begin{tcolorbox}[
    colframe=gray!50!black,
    colback=gray!10!white,
    title=Inpainting Selection Prompt,
    width=\textwidth, sharp corners=southwest,
    left=1mm, right=1mm, top=0mm, bottom=0mm
]
\scriptsize
\begin{lstlisting}[
    aboveskip=0pt,
    belowskip=0pt,
    breakatwhitespace=false,
    basicstyle=\ttfamily\scriptsize,
    showstringspaces=false,
    frame=none
]
You are picking the best inpainting result for a vector-graphic part.

INPUTS
- IMAGE 1 = original scene. The part labeled "{label}" exists somewhere in
  it but is partially OCCLUDED by other layers drawn on top.
- IMAGE 2 = the SAM cutout of the "{label}" layer on white. This is the
  partial / visible portion of the part (the rest is hidden in the scene).
- IMAGES 3..N = candidate inpainting results (A, B, C, D, ...) that
  attempted to reconstruct the FULL "{label}" by filling in the OCCLUDED
  region. The visible region was preserved from IMAGE 2 in all of them,
  so candidates differ ONLY in how they filled the occluded portion.

MOST IMPORTANT CRITERION (this dominates)
- Did the candidate COVER and DRAW the OCCLUDED region well? That is,
  the part of "{label}" that was missing in IMAGE 2 must be recovered
  as a sensible, complete shape that extends the visible silhouette
  naturally and FILLS the gap. A candidate that leaves the occluded
  region empty, broken, or only partially filled is worse than one
  that fully and plausibly reconstructs it. Pick the candidate that
  most COMPLETELY recovers the hidden portion of the "{label}".

Secondary criteria (only used to break ties between candidates that
already cover the occluded region equally well):
- It should not hallucinate unrelated objects/faces in the filled region.
- Color and edge style should roughly match the visible portion.
- It should not bleed far into regions that were obviously background.

OUTPUT -- JSON only, no markdown:
{{"best": "<A|B|C|D...>",
 "reason": "one short sentence naming the visible evidence (focus on occluded coverage)"}}

\end{lstlisting}
\end{tcolorbox}
\caption{\textbf{Inpainting Selection Prompt.}}
\label{fig:prompt_inpainting_selection}
\end{figure}

\vspace{0.5em}
\vspace{1em} 
\begin{tcolorbox}[
    breakable, 
    colframe=gray!50!black,
    colback=gray!10!white,
    title= Image-to-SVG Prompt for the General Purpose VLMs,
    width=\textwidth, sharp corners=southwest,
    left=1mm, right=1mm, top=0mm, bottom=0mm
]
\scriptsize
\begin{lstlisting}[
    aboveskip=0pt,
    belowskip=0pt,
    breakatwhitespace=false,
    basicstyle=\ttfamily\scriptsize,
    showstringspaces=false,
    frame=none
]

You are converting a raster image into a semantically structured, editable SVG file, exactly as a professional vector designer would author it.

Requirements:
1. Reproduce the image as faithfully as possible using flat-color vector paths.
2. Organize ALL paths into a semantic hierarchy of nested <g> tags. Every group must have a descriptive id (e.g. id="sun", id="cloud", id="mountain").
3. Order elements back-to-front in painter's order: background objects first, foreground objects last.
4. Each object must be GEOMETRICALLY COMPLETE (amodal): draw its full shape even where it is occluded by objects in front of it, so that removing or moving any foreground group leaves complete objects behind with no holes.
5. Use a viewBox that matches the image aspect ratio.

Output ONLY the final SVG code, nothing else.
\end{lstlisting}
\end{tcolorbox}
\captionof{figure}{\textbf{Image-to-SVG Prompt for the General Purpose VLMs.}}
\label{fig:prompt_grouping_svg}

\vspace{0.5em}
\begin{figure}[H]
\begin{tcolorbox}[
    colframe=gray!50!black,
    colback=gray!10!white,
    title=Text-to-SVG Prompt for the General Purpose VLMs,
    width=\textwidth, sharp corners=southwest,
    left=1mm, right=1mm, top=0mm, bottom=0mm
]
\scriptsize
\begin{lstlisting}[
    aboveskip=0pt,
    belowskip=0pt,
    breakatwhitespace=false,
    basicstyle=\ttfamily\scriptsize,
    showstringspaces=false,
    frame=none
]
You are creating a semantically structured, editable SVG file from the following description, exactly as a professional vector designer would author it.

Description: {text}

Requirements:
1. Depict the description as faithfully as possible using flat-color vector paths.
2. Organize ALL paths into a semantic hierarchy of nested <g> tags. Every group must have a descriptive id (e.g. id="sun", id="cloud", id="mountain").
3. Order elements back-to-front in painter's order: background objects first, foreground objects last.
4. Each object must be GEOMETRICALLY COMPLETE (amodal): draw its full shape even where it is occluded by objects in front of it, so that removing or moving any foreground group leaves complete objects behind with no holes.
5. Use a square viewBox.

Output ONLY the final SVG code, nothing else.

\end{lstlisting}
\end{tcolorbox}
\caption{\textbf{Text-to-SVG Prompt for the General Purpose VLMs.}}
\label{fig:prompt_text2svg}
\end{figure}

\clearpage

\twocolumn\flushbottom 

\end{document}